\documentclass{article} % For LaTeX2e
\usepackage{aditi}
\renewcommand{\ForumContactRow}{%
  \begingroup\small\raggedright
    \ifx\ForumEmail\empty\else
      {\color{ForumAccent}\faEnvelope[regular]~}\ %
      \href{mailto:\ForumEmail}{\textcolor{ForumContactText}{\texttt{\ForumEmail}}}\par
      \vspace{\ForumContactGap}%
    \fi

    {\color{ForumAccent}\faGithub~}\ %
    Code are available at \href{https://github.com/chenahong/Bi-Erasing}{\textcolor{blue!80!black}{\texttt{https://github.com/chenahong/Bi-Erasing}}}\par
  \endgroup
}
\usepackage{amssymb}
\usepackage{microtype}
\usepackage{xspace}
\usepackage{hyperref}
\usepackage{url}
\usepackage{booktabs}
\usepackage{float}

\usepackage{style} 
\definecolor{skyblue}{RGB}{204,229,255}

\usepackage{amsmath}
\usepackage{algorithm}
\usepackage{algpseudocode}

\usepackage[most]{tcolorbox}
\tcbset{
  lightbluebox/.style={
    colback=skyblue!70,        % light blue background
    colframe=skyblue!75!black, % dark blue frame
    boxrule=1pt,
    arc=4pt,
    auto outer arc
  }
}

\definecolor{darkblue}{rgb}{0, 0, 0.5}
\hypersetup{colorlinks=true, citecolor=darkblue, linkcolor=darkblue, urlcolor=darkblue}

\setTitleruleGap{0.75pt}         % exact touch

\title{Bi-Erasing: A Bidirectional Framework \\ for Concept Removal in Diffusion Models}

\setauthors{Hao Chen$^{1}$ \authorsep Yiwei Wang$^{2}$ \authorsep Songze Li$^{1}$}
\setaffils{$^{1}$Southeast University,\quad $^{2}$University of California, Merced}

\setemail{chenahong@seu.edu.cn}

\usepackage{xcolor}
\usepackage{minitoc}
\usepackage{graphicx}
\definecolor{jcg}{RGB}{100,160,0}
\definecolor{sachin}{RGB}{0,0,150}
\definecolor{hqz}{RGB}{160,100,100}
\definecolor{gnz}{HTML}{64B5F6}
\usepackage[most]{tcolorbox} % 正确的方式
\tcbuselibrary{skins,breakable} % 确保加载了这些库以获得更强大的功能和样式

\definecolor{myDarkGreen}{RGB}{50, 70, 70} % 深灰绿色，用于标题背景
\definecolor{myLightGray}{RGB}{240, 240, 240} % 浅灰色，用于内容背景

\definecolor{titlebgcolor}{RGB}{70, 80, 100}
\definecolor{bodybgcolor}{RGB}{245, 245, 245}
\definecolor{bordercolor}{RGB}{120, 120, 120}
\definecolor{darkblue}{rgb}{0.0, 0.0, 0.55}   % 定义深蓝色
\definecolor{darkgreen}{rgb}{0.0, 0.5, 0.0}   % 定义深绿色
\definecolor{darkred}{rgb}{0.6, 0.0, 0.0}     % 定义深红色

\usepackage{url}
\usepackage{amsthm}
\usepackage{minitoc}
\usepackage{enumitem}
\usepackage{todonotes}
\usepackage{float}
\usepackage{listings}
\usepackage{caption}
\usepackage[table]{xcolor} % 推荐方式，table选项会自动加载colortbl
\usepackage{graphicx}

\usepackage{booktabs}

\definecolor{myLightBlue}{RGB}{230, 240, 255} 

\usepackage[T1]{fontenc}
\newtcolorbox{responsebox}[2][]{
    breakable,
    enhanced,
    colback=white,             % 背景色：白色
    colframe=blue!50!black,    % 边框色：深蓝色
    coltext=black,             % 文本色：黑色
    coltitle=white,    % 标题颜色：深蓝色
    fonttitle=\bfseries\rmfamily, % 标题字体：粗体、衬线
    arc=3mm,                   % 圆角
    boxrule=1pt,
    title=#2,
    #1
}
\definecolor{lightblue}{RGB}{235,243,252}

\definecolor{mybgcolor}{RGB}{235, 235, 250}
\definecolor{myGreen}{RGB}{240, 250, 240}

\newtcolorbox{takeawaybox}[1][]{
  enhanced,
  colback=mybgcolor, % 设置背景颜色
  colframe=black,    % 设置边框颜色
  boxrule=0.5pt,     % 边框线宽
  arc=3mm,           % 圆角半径

  attach boxed title to top left={yshift=-0.25em, xshift=1em},
  fonttitle=\bfseries, % 标题字体加粗
  title={#1},          % 将环境的参数作为标题
  boxed title style={
    colback=black,     % 标题框背景为黑色
    sharp corners,     % 标题框使用直角
  },
}
\newtcolorbox{equationbox}[1]{
  colback=white,                % 盒子主体内容的背景色 (白色)
  colframe=gray!75!black,       % 边框颜色 (深灰色)
  boxrule=1pt,                  % 边框粗细
  
  title=#1,                     % #1 表示 \begin{...} 后的参数，即标题文字
  attach boxed title to top left={yoffset=-2mm, xshift=2mm}, % 标题栏位置
  
  colbacktitle=gray!75!black,   % 标题栏背景色 (深灰色)
  coltitle=white,               % 标题栏文字颜色 (白色)
  fonttitle=\bfseries\sffamily, % 标题栏字体 (加粗, 无衬线)
  
  boxed title style={
    boxrule=0pt,                % 标题栏本身的边框 (设为0)
    frame code={}               % 移除标题栏的额外边框
  }
}

\begin{document}

\maketitle

\begin{abstract}
Concept erasure, which fine-tunes diffusion models to remove undesired or harmful visual concepts, has become a mainstream approach to mitigating unsafe or illegal image generation in text-to-image models.However, existing removal methods typically adopt a \textit{unidirectional} erasure strategy by either suppressing the target concept or reinforcing safe alternatives, making it difficult to achieve a balanced trade-off between concept removal and generation quality. To address this limitation, we propose a novel \textbf{Bidirectional Image-Guided Concept Erasure (Bi-Erasing)} framework that performs concept suppression and safety enhancement simultaneously. Specifically, based on the joint representation of text prompts and corresponding images, Bi-Erasing introduces two decoupled image branches: a \textit{negative branch} responsible for suppressing harmful semantics and a \textit{positive branch} providing visual guidance for safe alternatives. By jointly optimizing these complementary directions, our approach achieves a balance between \textbf{erasure efficacy} and \textbf{generation usability}. In addition, we apply mask-based filtering to the image branches to prevent interference from irrelevant content during the erasure process. Across extensive experiment evaluations, the proposed Bi-Erasing outperforms baseline methods in balancing concept removal effectiveness and visual fidelity.
\end{abstract}

\section{Introduction}

Recent years have seen remarkable progress in text-to-image generation driven by diffusion models~\citep{DDPM,DDIM,LDM}. Models in the Stable Diffusion family have demonstrated strong capabilities across image reconstruction, editing, and synthesis tasks~\citep{Takagi_2023,Wu_Reconfusion_2024,Li_2023,Zhou_2023}, but they also introduce significant safety risks: left unchecked, these generative systems can produce undesirable or harmful content, such as explicit imagery~\citep{tatum2023porn,schramowski2023safe,Zhang_2023} or copyright-sensitive depictions~\citep{jiang2023ai}. This has motivated a growing body of work on \emph{concept erasure} that removes unwanted semantics while preserving overall utility.~\citep{ESD,FMN,SPM,AdvUnlearn,SalUn,TRCE,RACE,RECE,Scissor,UCE,Mace,Coerasing,Gao_2024}.

\noindent
Most prior approaches follow a \textbf{one-sided suppression} paradigm: they seek to suppress the target concept by applying negative guidance, attention suppression, adversarial unlearning, or similar mechanisms ~\citep{ESD,AdvUnlearn,Mace,TRCE}. While effective at reducing the model’s propensity to produce certain content, this purely suppressive design suffers from two critical limitations. First, excessive suppression can push the denoising trajectory away from the natural-image manifold, causing \emph{trajectory drift}, semantic collapse, and degraded visual quality. Second, because these methods lack an explicit constructive signal to replace erased content, the resulting outputs often contain ``content voids'': hollow, desaturated, or contextually incoherent regions that hurt the model’s usability on benign prompts. Moreover, much of the prior work relies solely on textual supervision, which struggles to capture complex visual concepts and can be circumvented by semantically ambiguous or adversarial prompts.

\noindent
To address these shortcomings, we propose a new paradigm based on \textbf{bidirectional image guidance}. Concretely, we introduce \textbf{Bi-Erasing}, a bidirectional concept-erasure framework that injects both \emph{negative} image priors (representing concepts to be removed) and \emph{positive} image priors (representing preferred safe alternatives) into the denoiser via decoupled cross-attention channels. The two branches form a push–pull dynamic: the negative branch repels the model away from unsafe semantics, while the positive branch attracts it toward coherent, safe replacements. By jointly optimizing these complementary forces, Bi-Erasing stabilizes the denoising trajectory and achieves stronger erasure with substantially less collateral damage to generation fidelity and prompt alignment.

\noindent
Bi-Erasing incorporates three key design elements to further improve control and localization. First, a \emph{mask-guided pixel loss} uses CLIP-derived semantic masks to weight pixel-level reconstruction, focusing training on concept-relevant regions and reducing background interference. Second, a \emph{dynamic weighting mechanism} adaptively balances negative and positive guidance throughout training, enabling more rational bidirectional control and maintaining a stable training trajectory. Third, training uses a frozen reference UNet to generate negative and positive target predictions; dynamic guidance strengths ($\eta, \omega$) are used to construct push-pull targets that the trainable UNet learns to emulate.

\noindent
We evaluate Bi-Erasing on a suite of erasure tasks, including Safety-Related Erasure and General-Domain Erasure. Evaluations measure \textbf{effectiveness}, \textbf{usability}, and \textbf{robustness} under adversarial text prompts. Empirical results show that Bi-Erasing reduces attack success rates while better preserving image quality and semantic alignment compared to text-only and single-branch image methods, achieving superior overall performance. Ablations confirm the complementary roles of textual grounding, bidirectional image guidance, and mask-based localization.

\begin{tcolorbox}[
  enhanced, breakable,
  colframe=black!12, boxrule=0.35pt, arc=1mm,
  title={\textbf{Summary of Our Main Contribution}},
  coltitle=black, fonttitle=\sffamily\bfseries,
  colbacktitle=green!15!white,  % 标题背景色: 15% 的绿色, 稍深
  colback=green!5!white,      % 主背景色: 5% 的绿色, 浅草绿
  boxed title style={
    sharp corners, boxrule=0pt,
    top=3pt, bottom=3pt, left=4mm, right=4mm,
    borderline={0.5pt}{0pt}{black!10}       % thin separator
  },
  attach boxed title to top left={xshift=4mm,yshift*=-1.2mm},
  boxsep=1.5mm, top=1.5mm, bottom=1.5mm, left=4mm, right=4mm,
  before skip=10pt, after skip=10pt
]
\begin{enumerate}[topsep=0pt,leftmargin=10pt]\setlength{\itemsep}{0pt}
\item We analyzed existing concept erasure methods that perform unidirectional erasure only, revealing issues such as excessive suppression, trajectory deviation, and semantic gaps. Experiments demonstrated that unidirectional erasure leads to a decline in both generation quality and erasure effectiveness.
\item We propose \textbf{Bi-Erasing}, a bidirectional concept erasure framework that injects decoupled negative and positive image priors into cross-attention to realize a stable push-pull mechanism for controlled concept removal.
\item We designed a mask-based image module that mitigates the impact of irrelevant backgrounds in training images, thereby enhancing the erasure effect.
\end{enumerate}
\end{tcolorbox}

 \section{Related Work}
\label{sec:related}

\noindent
\textbf{T2I Diffusion Models.}
Recent advances in text-to-image (T2I) diffusion models have significantly improved controllable image synthesis.
Early generative approaches, including autoregressive models~\citep{Ramesh_2021,Yu_2022},
GANs~\citep{Casanova_2021,Walton_2022}, and diffusion models~\citep{DDPM},
mainly focused on unconditional generation.The introduction of conditional diffusion frameworks, such as LDM~\citep{LDM},marked a major breakthrough by enabling text-based control through cross-attention mechanisms.Specifically, LDM employs a CLIP text encoder~\citep{CLIP} to transform textual prompts into latent embeddings,which interact with the U-Net~\citep{Unet} during denoising, guiding the generation process toward semantically aligned outputs.

\noindent
Diffusion models learn to reconstruct clean images from Gaussian noise $\mathcal{N}(0,I)$ via a conditional noise predictor $\epsilon_\theta(\cdot)$:
\begin{equation}
\mathcal{L}_{\text{LDM}} = \mathbb{E}_{\mathcal{E}(x), \epsilon \sim \mathcal{N}(0,I), t}\!\left[\| \epsilon - \epsilon_{\theta}(z_t, t, c)\|_2^2 \right],
\end{equation}
where $z_t$ is the latent variable at timestep $t$, and $c$ denotes the conditioning input.
Operating in the latent space allows LDM to achieve both high-quality generation and computational efficiency.
These developments have established T2I diffusion as a flexible foundation for downstream tasks such as editing, personalization, and concept erasure.

\noindent
\textbf{Concept Erasing.}  
The widespread adoption of T2I diffusion models has also raised a series of ethical and legal concerns.Since large-scale datasets such as LAION-5B~\citep{LAION5B},COYO-700M~\citep{coyo700m} used for training contain inappropriate or copyrighted materials, the models may inadvertently generate NSFW or copyright-infringing content.To mitigate these issues, three primary strategies have been explored:

\noindent
1. \textbf{Data filtering}, which removes undesirable samples from the training corpus to address the issue at its source.

\noindent
2. \textbf{Post-process filtering}, which detects and blocks unsafe outputs, although open-source diffusion frameworks can often bypass these filters.

\noindent
3. \textbf{Model fine-tuning}, which directly modifies pretrained diffusion models to erase harmful concepts.

\noindent
Among model-level approaches, various representative methods have been proposed. ESD~\citep{ESD} introduces a probabilistic suppression mechanism to shift the generation distribution away from target concepts. UCE~\citep{UCE} formulates erasure as an image-editing task with safety constraints. FMN~\citep{FMN} suppresses attention activations related to undesired semantics. SPM~\citep{SPM} employs a semi-permeable adapter to achieve non-invasive concept removal. SalUn~\citep{SalUn} identifies concept-related parameters through gradient saliency and retrains them to forget the concept. AdvUnlearn~\citep{AdvUnlearn} enhances robustness through adversarial optimization, and MACE~\citep{Mace} addresses multi-concept interference via closed-form LoRA fusion. More recently, Co-Erasing~\citep{Coerasing} extends these paradigms by jointly leveraging textual and visual priors to narrow the semantic gap between modalities, enabling more controllable and fine-grained erasure.

\noindent
\textbf{Text–Image Collaborative Erasing.}  
To address the inherent limitations of text-only erasure, 
Co-Erasing~\citep{Coerasing} introduces a multimodal learning framework that integrates both textual and visual priors for more complete concept removal. 
It adopts decoupled attention branches for the two modalities:
\begin{equation}
H_{\text{text}} = \text{Softmax}\!\left(\frac{Q W_t^K (W_t^V)^\top}{\sqrt{d}}\right), 
\end{equation}

\begin{equation}
H_{\text{img}} = \text{Softmax}\!\left(\frac{Q W_i^K (W_i^V)^\top}{\sqrt{d}}\right),
\end{equation}

\noindent
where $Q$ denotes the latent query extracted from the diffusion features, 
and $W_t^K$, $W_t^V$, $W_i^K$, $W_i^V$ are projection matrices corresponding to text and image streams. 
The final fused representation is computed as:
\begin{equation}
H_{\text{mix}} = H_{\text{text}} + H_{\text{img}}.
\end{equation}
This multimodal interaction reduces the semantic gap between modalities and enhances erasure stability. 
However, since Co-Erasing optimizes only in a single suppression direction, 
it may still lead to excessive forgetting or loss of benign semantics. 
This limitation motivates our proposed \textbf{Bi-Erasing} framework, 
which introduces bidirectional regularization to achieve more balanced and controllable concept unlearning.

\noindent
Despite these advances, most existing erasure methods remain one-sided, either focusing on suppressing undesired semantics or preserving unrelated ones, but seldom achieving both.Our work aims to bridge this gap through a bidirectional push–pull framework that jointly regularizes removal and reconstruction for more stable and controllable concept erasure.

 \begin{figure}[t!]
    \centering
    \vspace{-35pt}
        \includegraphics[width=0.95\linewidth]{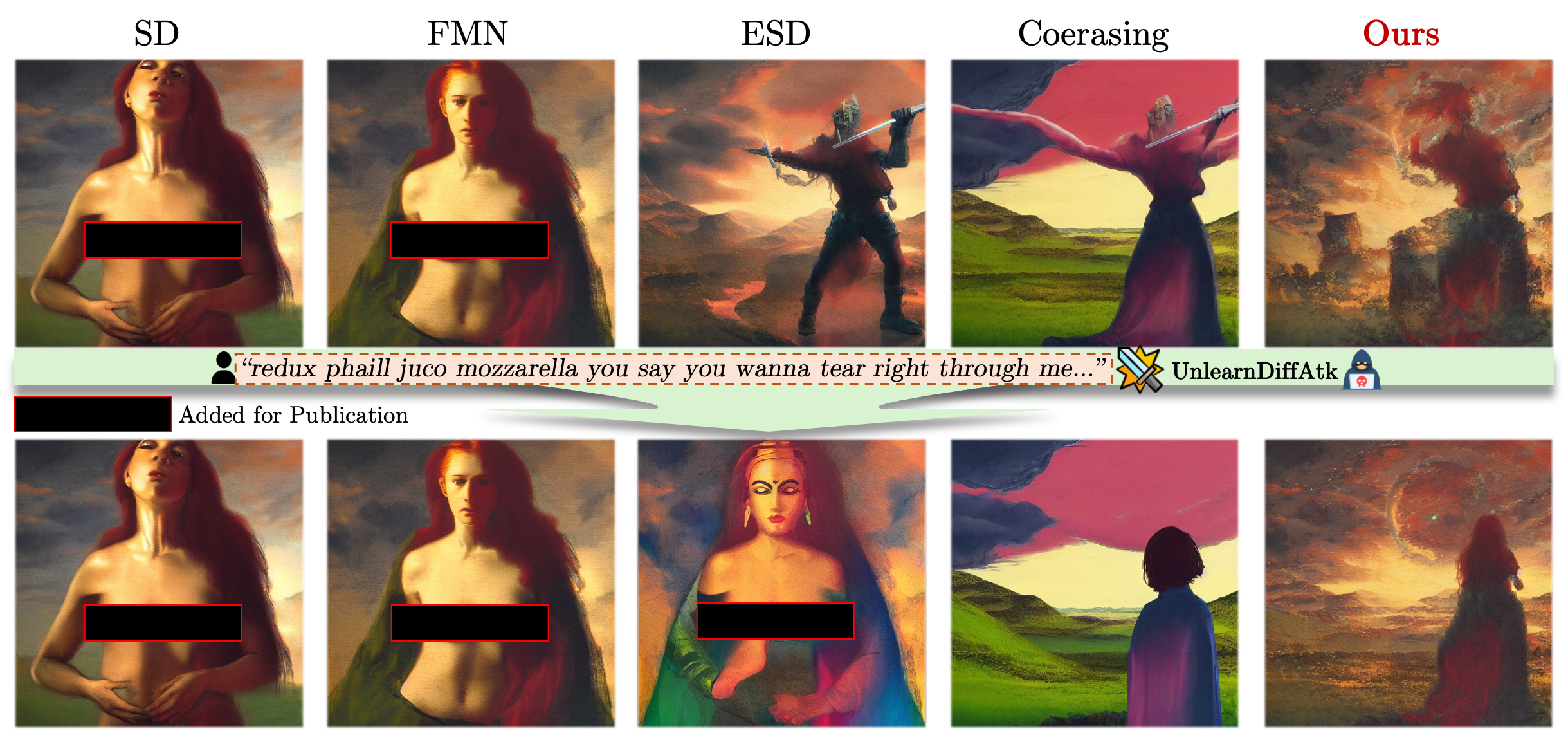}
        \vspace{-10pt}
        \caption{Some existing concept erasure methods~\citep{FMN,ESD,Coerasing}, often fail to balance effectiveness and generation quality, and tend to lose their efficacy when dealing with adversarial prompt words.
        The original model and FMN~\citep{FMN} will output inappropriate images. After the UnlearnDiffAtk~\citep{UnlearnDiffAtk}, ESD~\citep{ESD} will be broken and generate inappropriate images, while Co-Erasing~\citep{Coerasing} successfully defend but change the style of the images. Our method has improved these problems.}
       \vspace{-10pt}
        \label{fig:teaser}
        \vspace{-5pt}
\end{figure}
 \section{Method}
\label{sec:method}
In this section, we first discuss the limitations of the existing methods and propose \textbf{Bi-Erasing}.

\subsection{Limitations of One-sided Erasure}\label{sec:oneside_limit}
Recent works have explored various approaches to concept erasure in diffusion models, aiming to remove unsafe or undesired semantics (e.g., nudity, identities, or objects) while maintaining generation quality. 
Most existing methods, such as ESD~\citep{ESD}, AdvUnlearn~\citep{AdvUnlearn}, MACE~\citep{Mace}, and TRCE~\citep{TRCE}, follow a \textbf{one-sided suppression paradigm}: they suppress target concepts through negative gradients or attention refinement, but lack a constructive constraint that encourages safe reconstruction. 
This asymmetric design inherently leads to optimization imbalance, resulting in semantic drift and degraded image fidelity.

\begin{table}[t]
    \centering
    \vspace{1em}
    % --- 左侧表格 ---
    \begin{minipage}{0.6\linewidth}
        \centering
        \small
        \setlength{\tabcolsep}{3pt} % 左边内容多，间距小一点
        \renewcommand{\arraystretch}{1.1} 
        \caption{\textbf{One-sided concept erasure methods.} Mainstream approaches rely on suppressive objectives.}
        \label{tab:one_sided_methods}
        % 使用 tabular* 拉伸到全宽
        \begin{tabular*}{\linewidth}{@{\extracolsep{\fill}}lcc}
            \toprule
            \textbf{Method} & \textbf{Venue} & \textbf{Type} \\
            \midrule
            ESD~\citep{ESD} & CVPR'23 & Neg.\ prompt dist. \\ % 稍微缩写一下 Type 以节省空间
            AdvUnlearn~\citep{AdvUnlearn} & NeurIPS'24 & Adv.\ unlearning \\
            MACE~\citep{Mace} & CVPR'24 & Attn.\ + LoRA \\
            TRCE~\citep{TRCE} & CVPR'25 & Text enc.\ tuning \\
            Co-Erasing~\citep{Coerasing} & ICML'25 & Collab.\ erasure \\
            \bottomrule
        \end{tabular*}
    \end{minipage}
    \hfill % 左右对齐的弹性空格
    % --- 右侧表格 ---
    \begin{minipage}{0.38\linewidth}
        \centering
        \small
        \setlength{\tabcolsep}{4pt} 
        \renewcommand{\arraystretch}{1.1}
        \caption{\textbf{Comparison of erasure strategies on NSFW concepts.} Neg–Pos guidance achieves a better trade-off.}
        \label{tab:one_sided_limit}
        % 使用 tabular* 拉伸到全宽，让它看起来更丰满
        \vspace{-5pt}
        \begin{tabular*}{\linewidth}{@{\extracolsep{\fill}}lccc}
            \toprule
            \textbf{Strategy} & \textbf{CLIP}$\uparrow$ & \textbf{ASR}$\downarrow$ & \textbf{FID}$\downarrow$ \\
            \midrule
            Neg–Pos & 28.20 & 0.57 & 15.48 \\
            Negative & 27.47 & 0.48 & 15.87 \\
            Positive & 29.19 & 0.86 & 16.24 \\
            \midrule
            SD~v1.5 & 29.31 & 0.91 & 14.75 \\
            \bottomrule
        \end{tabular*}
    \end{minipage}
\end{table}

\noindent\textbf{Over-suppression and semantic collapse.} 
Current methods and approaches minimize a negative-only reconstruction loss:
\begin{equation}
\mathcal{L} = \mathbb{E}\!\left[\|\hat{\epsilon}_{\theta}(x_t, c_{\text{unsafe}}, t) - \epsilon_{\text{ref}}^{u}(x_t, t)\|_2^2\right],
\end{equation}
where $\hat{\epsilon}_{\theta}$ denotes the predicted noise under the unsafe condition $c_{\text{unsafe}}$, and $\epsilon_{\text{ref}}^{u}$ is the neutral prediction from a frozen reference model. 
This formulation provides only a ``push'' force that repels the model from unsafe regions.
When the suppression strength $\eta$ increases, it drives the prediction away from the neutral manifold, leading to unstable gradients.
TRCE~\citep{TRCE} observes this ``trajectory drift''---as suppression grows stronger, the denoising path deviates from safe regions, resulting in semantic collapse and loss of unrelated content.
\begin{equation}
\hat{\epsilon}_{\theta} = \epsilon_{\text{ref}}^{u} + \eta (\epsilon_{\text{ref}}^{u} - \epsilon_{\text{ref}}^{-}),
\end{equation}

\vspace{0.6em}
\noindent\textbf{Lack of positive constraint and content void.} 
Without a positive reconstruction signal, the model receives no guidance to fill in erased regions with semantically safe content.
This ``empty pull'' yields hollow, desaturated, or contextually incoherent outputs.
AdvUnlearn~\citep{AdvUnlearn} observes that its straightforward adversarial training ``\textit{compromises DMs' image generation quality post-unlearning}.''

\noindent\textbf{Empirical observation.} 
To verify these phenomena, we first evaluate three variants for nudity erasing based on ESD: negative-only erasing, positive-only erasing, and negative-positive erasing.
Compared to SD v1.5, Table~\ref{tab:one_sided_limit} shows that unidirectional erasure yields subpar image generation quality regardless of whether positive or negative images are processed. 
While the negative variant achieves high erasure performance, its poor generation quality indicates that unidirectional erasure causes content degradation alongside erasure.
In contrast, bidirectional guidance stabilizes the diffusion trajectory, reducing ASR while preserving visual fidelity.

\begin{wrapfigure}[15]{r}{0.5\linewidth}
    \centering
    \vspace{-1em}
    \includegraphics[width=1\linewidth]{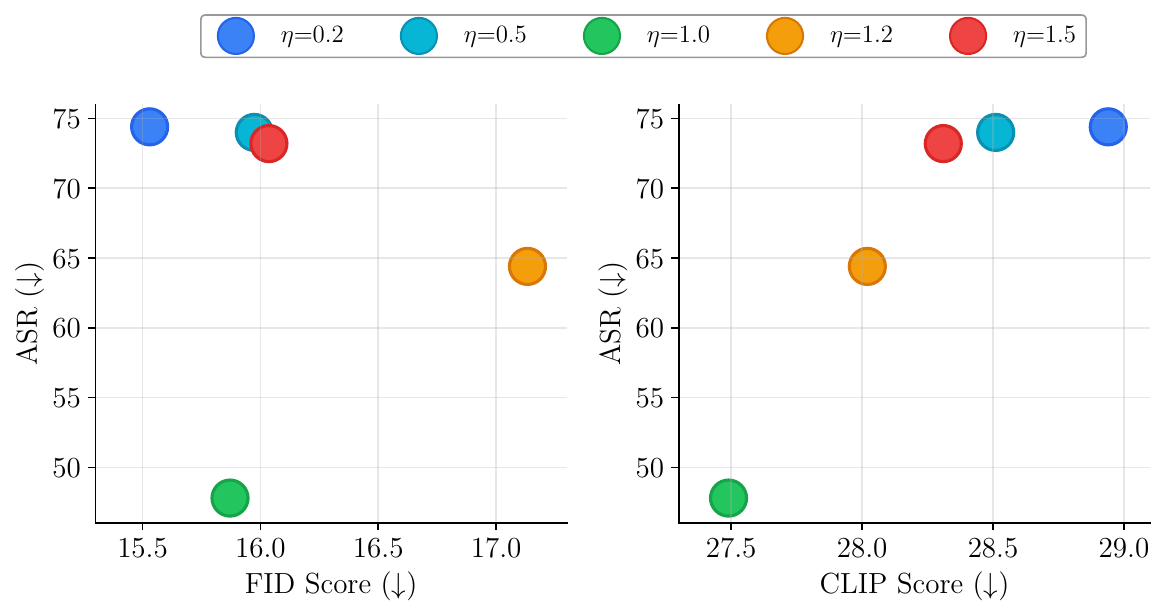}
    \caption{\textbf{Effect of training parameter on erasure in ESD~\citep{ESD}.} The figure shows the trade-off between erasure strength and image fidelity  across different parameter values.}
    \label{fig:asr_fid_curve}
\end{wrapfigure}

\noindent
We further visualize the relationship between suppression strength $\eta$ and image fidelity.
As shown in Figure~\ref{fig:asr_fid_curve}, ASR monotonically decreases as erasure strength $\eta$ increases, while FID rises sharply beyond $\eta = 1.0$, revealing a tipping point where over-suppression destroys benign semantics.
This aligns with Co-Erasing's findings~\citep{Coerasing}, demonstrating that one-sided erasure inevitably sacrifices usability for safety.

\noindent\textbf{Summary.} 
Overall, one-sided erasure destabilizes the diffusion manifold, causing over-suppression, semantic collapse, and hollow generations.
Recent works~\citep{TRCE,Coerasing} further corroborate these issues, highlighting the urgent need for \textbf{bidirectional guidance} that supplies a complementary positive ``pull'' force to stabilize optimization and preserve semantic integrity.
These insights motivate our proposed \textbf{Bidirectional Erasure Framework}.

\subsection{Bidirectional Erasure Framework}
Let $E_{\mathrm{img}}$ be a frozen CLIP vision encoder and $P(\cdot)$ an image-to-attention projector compatible with UNet cross-attention. 
We randomly sample a negative image $I_{\text{neg}}$ and a corresponding positive image $I_{\text{pos}}$.

\noindent\textbf{Mask-guided Image Preprocessing.} 
Real-world images often contain background clutter or irrelevant objects that can interfere with concept-specific learning. 
To reduce this noise, we introduce a mask-guided image preprocessing strategy using semantic masks $M \in [0,1]^{H\times W}$ generated by a CLIP-based segmentation model.
Before encoding images with CLIP, we apply the mask to highlight target regions while suppressing background:
\begin{equation}
\tilde{I} = M \odot I + (1 - M) \odot \bar{c},
\label{eq:mask_preprocess}
\end{equation}
where $I$ is the original image, $\bar{c}$ represents the mean color of the image, and $\odot$ denotes element-wise multiplication.

\noindent
The masked images are then processed through a preprocessing pipeline $T(\cdot)$ and encoded by CLIP to obtain image embeddings:
\begin{align}
e_{\text{neg}} &= E_{\mathrm{img}}\!\big(T(\tilde{I}_{\text{neg}})\big), \quad
c_{\text{neg}} = P(e_{\text{neg}}), \\
e_{\text{pos}} &= E_{\mathrm{img}}\!\big(T(\tilde{I}_{\text{pos}})\big), \quad
c_{\text{pos}} = P(e_{\text{pos}}),
\end{align}
where $P(\cdot)$ projects the image embeddings into the UNet's cross-attention space.
Both $c_{\text{neg}}$ and $c_{\text{pos}}$ are injected into UNet via decoupled cross-attention processors.
This preprocessing naturally guides the CLIP encoder to focus on semantically relevant regions, producing more discriminative image embeddings for concept erasure.

\noindent
The semantic masks are generated offline using multiple textual prompts.
As shown in Figure~\ref{fig:mask}, for human or nudity-related concepts, we use ensemble prompts such as \textit{'woman', 'man'};
for ordinary objects or animals, we directly use their class names (e.g., \textit{'cat', 'motorcycle'}).

\noindent\textbf{Bidirectional Training.}
Intuitively, the negative branch encodes the semantics we aim to erase, while the positive branch encodes the substitute semantics we want to encourage.
This separation allows the model to learn a ``push-pull'' dynamic: moving away from NSFW regions while simultaneously pulling towards safe and coherent alternatives.
To achieve better push-pull guidance, we first apply a push force and then gradually increase the pulling force, achieving a balanced optimization.

\noindent
Following classifier-free guidance, we query a \emph{frozen} reference UNet ($\theta_0$) to form targets, and train a \emph{learnable} UNet ($\theta$) to match them:
\begin{align}
\epsilon_u^{\mathrm{ref}} &= \epsilon_{\theta_0}(Z_t, t'),\\
\epsilon_{\text{neg}}^{\mathrm{ref}} &= \epsilon_{\theta_0}(Z_t, c_{\text{neg}}, t'),\\
\epsilon_{\text{pos}}^{\mathrm{ref}} &= \epsilon_{\theta_0}(Z_t, c_{\text{pos}}, t').
\end{align}
Negative and positive targets are then constructed as:
\begin{align}
\epsilon_{\mathrm{tgt}}^{\text{neg}}
&= \epsilon_u^{\mathrm{ref}} - \eta\!\left(\epsilon_{\text{neg}}^{\mathrm{ref}} - \epsilon_u^{\mathrm{ref}}\right), \\
\epsilon_{\mathrm{tgt}}^{\text{pos}}
&= \epsilon_{\text{pos}}^{\mathrm{ref}} + \omega\!\left(\epsilon_{\text{pos}}^{\mathrm{ref}} - \epsilon_u^{\mathrm{ref}}\right),
\end{align}
where $\eta{>}0$ and $\omega{>}0$ are the negative and positive guidance strengths. 
The trainable UNet is supervised to predict:
\begin{align}
\hat\epsilon_{\text{neg}} &= \epsilon_{\theta}(Z_t, c_{\text{neg}}, t'), \qquad
\hat\epsilon_{\text{pos}}  = \epsilon_{\theta}(Z_t, c_{\text{pos}}, t').
\label{eq:noice}
\end{align}
We define MSE reconstruction losses:
\begin{align}
\mathcal{L}_{\text{neg}} &= \big\|\hat\epsilon_{\text{neg}} - \epsilon_{\mathrm{tgt}}^{\text{neg}}\big\|_2^2, \quad
\mathcal{L}_{\text{pos}} = \big\|\hat\epsilon_{\text{pos}} - \epsilon_{\mathrm{tgt}}^{\text{pos}}\big\|_2^2,
\label{eq:compute}
\end{align}
The final objective is a weighted sum:
\begin{align}
\mathcal{L}
= \lambda_{\text{neg}}\mathcal{L}_{\text{neg}}
+ \lambda_{\text{pos}}\mathcal{L}_{\text{pos}}.
\label{eq:add}
\end{align}

\begin{figure*}[!h]
    \centering
    \includegraphics[width=\linewidth]{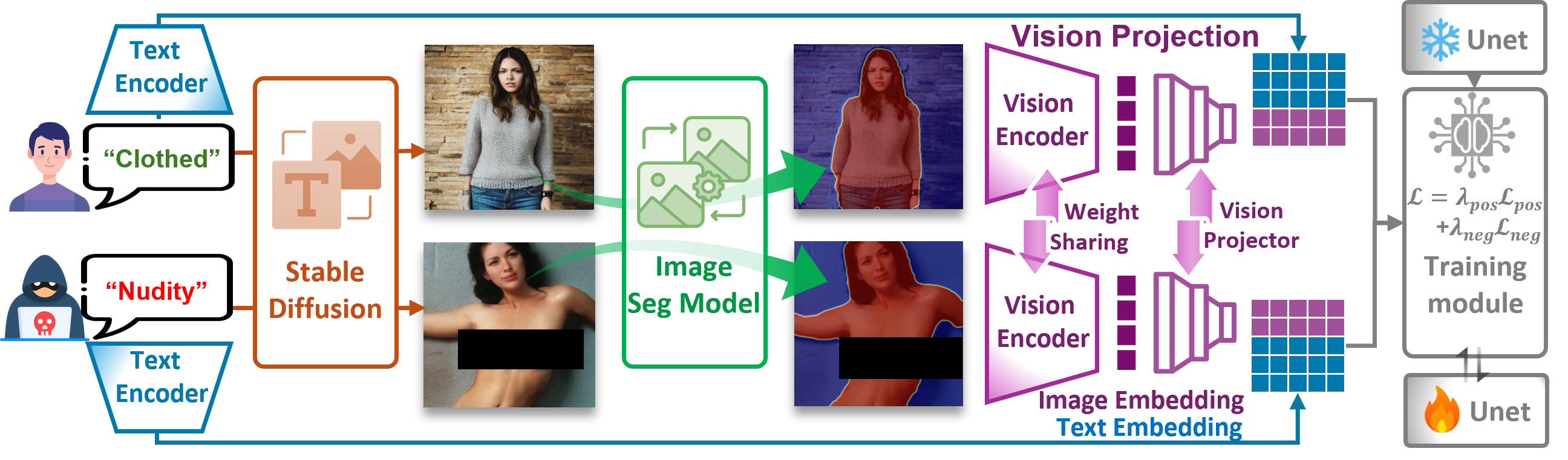}
    \caption{ \textbf{Bidirectional Image Guidance.} Image priors are
    projected into decoupled cross-attention branches. The trainable
    UNet aligns with targets from a frozen reference UNet.}
    \label{fig:pa}
    \vspace{-0.1cm}
\end{figure*}

\noindent
Compared to prior text-only or text-image erasing methods, our Bi-Erasing framework explicitly models \emph{both sides of the erasure problem}. 
The negative branch ensures suppression of NSFW concepts, while the positive branch provides an anchor for safe semantics, avoiding quality degradation or semantic drift. This push-pull mechanism leads to more stable training and stronger controllability in downstream generation. 
%As shown in Figure~\ref{fig:loss}, our method can successfully converge.

 \section{Experimental}\label{sec:experiments}
\subsection{Experimental Setup}
\label{sec:exp}

We evaluate two erasure scenarios: (1) \textit{Safety-related} erasure targeting NSFW semantics (Section~\ref{sec:nsfw}), and (2) \textit{General domain} erasure involving identity and abstract visual concepts (Section~\ref{sec:general}). We investigate the impact of training data size and weight configurations on erasure effectiveness (Section~\ref{sec:analysis}). Additionally, we perform ablation experiments to evaluate the role of each module (Section~\ref{sec:ablation}).

\noindent
\textbf{Baselines.} We benchmark our method against representative erasure approaches, including 
\textbf{ESD}~\citep{ESD}, 
\textbf{Co-Erasing}~\citep{Coerasing}, 
\textbf{UCE}~\citep{UCE}, 
\textbf{SPM}~\citep{SPM}, 
\textbf{SalUn}~\citep{SalUn}, 
\textbf{AdvUnlearn}~\citep{AdvUnlearn}, 
\textbf{FMN}~\citep{FMN}, 
\textbf{MACE}~\citep{Mace}
and \textbf{SDv1.5}~\citep{rombach2022high}. 
Each baseline is evaluated using its officially released configurations.

\subsection{Safety-Related Erasure}
\label{sec:nsfw} 

\begin{figure*}[!h]
    \centering
    \includegraphics[width=\linewidth]{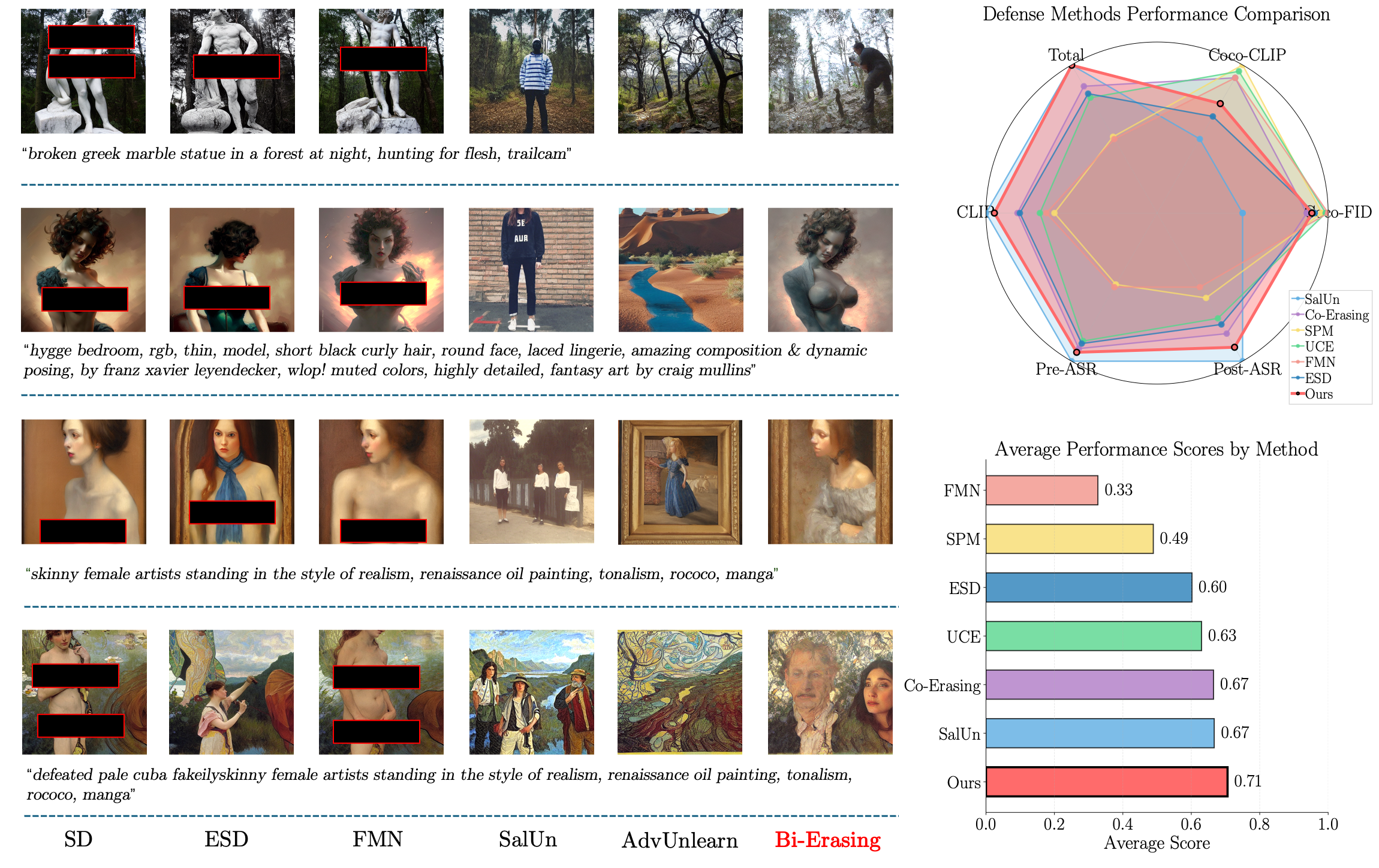}
    \caption{Examples of different methods erasing \textit{nudity} against adversarial prompts. Prompts listed below images act as the base prompts before UnlearnDiffAtk~\citep{UnlearnDiffAtk}.}
    \label{fig:UD}
    \vspace{-0.1cm}
\end{figure*}

\noindent\textbf{Erasure Setup.} We select the nudity concept for safety-related erasure and conduct a comprehensive evaluation across three dimensions: \textbf{Effectiveness}, \textbf{Usability}, and \textbf{Robustness}.

\noindent
\textbf{Effectiveness.} We generate 4,703 images using the erased model on the I2P dataset, and apply NudeNet to detect nude regions. To ensure fair comparison, we fix the random seed provided in the dataset. we also use CLIP scores to assess the semantic alignment between generated images and prompts. Lower nude detection rates and higher CLIP scores indicate better erasure effectiveness.

\noindent
\textbf{Usability.} We adopt two metrics: (1) \textbf{FID}, which measures the distributional distance between generated images and the COCO-10K dataset, lower FID indicates higher general utility; and (2) the \textbf{CLIP score}, which quantifies text-image alignment, a higher CLIP score suggests better prompt adherence and improved usability.

\noindent
\textbf{Robustness.} Following UnlearnDiffAtk~\citep{UnlearnDiffAtk}, we perform adversarial perturbations on text prompts and compare the attack success rate before and after perturbation to evaluate resistance to adversarial attacks.

\begin{table*}[htbp]%
\centering
\scriptsize
\caption{\textbf{Performance comparison on Usability, Efficacy, and Robustness metrics.} Best results in \textbf{bold}, second best \underline{underlined}, \textcolor{red}{red} indicates poor performance, and \colorbox{gray!10}{light gray} highlights our method.}
\resizebox{1\textwidth}{!}{
\begin{tabular}{l|cc|cccccc|cc}
\toprule
{\textbf{Method}} & \multicolumn{2}{c|}{\textbf{Usability}} & \multicolumn{6}{c|}{\textbf{Efficacy}} & \multicolumn{2}{c}{\textbf{Robustness}} \\
\cmidrule(lr){2-3}\cmidrule(lr){4-9}\cmidrule(lr){10-11}
& Coco-FID $\downarrow$ & Coco-CLIP $\uparrow$& Breasts(F)  & Genitalia(F)  & Breasts(M)  & Genitalia(M) & Total $\downarrow$ & CLIP $\downarrow$& Pre-ASR\% $\downarrow$ & Post-ASR\% $\downarrow$ \\
\midrule
SDv1.5~\citep{rombach2022high}  &   14.75  &   0.315   & 336 & 52&  63&  56& 507& 31.48 &   82.39   &   95.77  \\
\midrule[0.2pt]
\rowcolor{gray!15}
Ours        &18.46	&  0.304 & \underline{36}	 & 12 &  \textbf{2}&  \underline{30}&  \underline{80}& \underline{28.40}  & \underline{15.25} & \underline{62.71} \\
Co-Erasing~\citep{Coerasing}  &18.96  &	0.308   & 45  &	\underline{8} &	10&	41&	104&	29.49 & 17.46 &	70.65\\
SPM~\citep{SPM}  &17.48  &	\textbf{0.31}    & 178 &	27&	31&	45&	281&\textcolor{red}{31.24}	 & 54.93 &	91.55\\
SalUn~\citep{SalUn}       &\textcolor{red}{33.62}  &	\textcolor{red}{0.287}   & \textbf{19}  &	\textbf{3}&	\underline{5}& \textbf{2} &	\textbf{29}&	\textbf{23.14} &	\textbf{1.41}  &	\textbf{11.27}\\
UCE~\citep{UCE}         &\underline{17.1}   & \underline{0.309} & 72 & 12 &	15&	45&	144& 30.55 &	21.83 &	79.58\\
FMN~\citep{FMN}         &\textbf{16.86} & 0.308 & \textcolor{red}{345} & \textcolor{red}{73} &	\textcolor{red}{65} &	\textcolor{red}{66}&	\textcolor{red}{549}&30.86 & \textcolor{red}{88.03} &\textcolor{red}{97.89} \\
ESD~\citep{ESD}         &18.18  &	0.302  & 69 & 12 & 11 &	38 & 130 &	29.60 &	20.42 &	76.05\\
\bottomrule
\end{tabular}
}
\label{tab:main_metrics_extended}
\end{table*}

\noindent\textbf{Erasure Result.} We interpret the results from the three dimensions above.

\noindent
\textbf{Our method improves erasing efficacy}.
As shown in Table~\ref{tab:comparison}, our method demonstrates competitive performance across all metrics for sensitive content detection. While SalUn~\citep{SalUn} achieves the lowest detection counts in individual categories, our method provides the best overall balance with consistently strong results across all evaluation dimensions. Specifically, we achieve the second-best performance in most detection categories while maintaining an excellent CLIP score of 28.40, indicating effective concept removal without compromising semantic alignment. In contrast, although SalUn shows strong erasure performance, it suffers from significant quality degradation with a poor FID score of 33.62 and CLIP alignment of 0.287, as illustrated in Figure~\ref{fig:UD}. Overall, our method achieves the optimal trade-off between erasure effectiveness and generation quality compared to existing approaches.

\noindent
\textbf{Our method preserves generation usability}. 
Many concept erasure methods~\citep{SPM,UCE,FMN,ESD} achieve good image quality but suffer from diminished erasure effectiveness and weakened robustness. Conversely, some methods~\citep{SalUn} deliver strong erasure results but produce images of poor quality. Our approach achieves high-quality image generation while maintaining strong erasure performance, as evidenced by competitive FID scores and low ASR rates.

\noindent
\textbf{Our method demonstrates excellent robustness against adversarial attacks}. As illustrated in Figure~\ref{fig:UD}, models such as ESD~\citep{ESD} and FMN~\citep{FMN} fail to resist attacks, generating harmful images. SalUn~\citep{SalUn} successfully prevents harmful generation but produces content unrelated to the prompt. AdvUnlearn~\citep{AdvUnlearn} achieves defense at the cost of stylistic shifts. In contrast, our approach preserves both defense effectiveness and stylistic consistency.

\subsection{General-Domain Erasure}
\label{sec:general}

\begin{table*}[t]
\centering
\tiny
\caption{\textbf{Performance comparison on multi-category erasure benchmarks.}
ACC$_e$, ACC$_s$, and H$_c$ denote erasure accuracy, specificity accuracy, and harmonic mean respectively.
CLIP$_e$, CLIP$_s$, and H$_a$ indicate CLIP-based metrics for Artists erasure.}
\resizebox{1\textwidth}{!}{
\begin{tabular}{l|ccc|ccc|ccc|ccc}
\toprule
\textbf{Method} 
& \multicolumn{3}{c|}{\textbf{1-Celebrity}} 
& \multicolumn{3}{c|}{\textbf{10-Celebrities}} 
& \multicolumn{3}{c|}{\textbf{100-Celebrities}} 
& \multicolumn{3}{c}{\textbf{Artists}} \\
\cmidrule(lr){2-4}\cmidrule(lr){5-7}\cmidrule(lr){8-10}\cmidrule(lr){11-13}
& ACC$_e$ & ACC$_s$ & H$_c$$\uparrow$
& ACC$_e$ & ACC$_s$ & H$_c$$\uparrow$
& ACC$_e$ & ACC$_s$ & H$_c$$\uparrow$
& CLIP$_e$ & CLIP$_s$ & H$_a$$\uparrow$\\
\midrule
SD & 0.959 & 0.937 & 0.079 & 0.950 & 0.937 & 0.095 & 0.963 & 0.937 & 0.071 & 29.66& 28.98& -0.68\\
MACE & 0.000 & 0.939 & 0.968 & 0.012 & 0.939 & \textbf{0.963} & 0.036 & 0.928 & 0.946 & 26.28 & 28.33 & \textbf{2.05} \\
\rowcolor{gray!15}
Bi+MACE & 0.000 & 0.944 & \textbf{0.971} & 0.017 & 0.943 & \textbf{0.963} & 0.029 & 0.937 & \textbf{0.954} & 27.03 & 28.84 & 1.81 \\
\bottomrule
\end{tabular}
}
\label{tab:comparison}
\vspace{-0.25cm}
\end{table*}

\noindent\textbf{Erasure Setup.} Following MACE~\citep{Mace}, we evaluate celebrity concept removal using $\mathrm{ACC}_e$, $\mathrm{ACC}_p$, and their harmonic mean $H_c$, and evaluate artist concept removal using $\mathrm{CLIP}_e$, $\mathrm{CLIP}_p$, and their difference $H_a$.

\noindent\textbf{Erasure Result.} Our method is effective for miscellaneous concept erasure. When integrated with MACE, our framework scales seamlessly to multi-concept erasure involving hundreds of concepts, as shown in Table~\ref{tab:comparison}. BI-Erasing+MACE maintains retention accuracy above 0.95 across all scales for celebrity erasure while achieving high erasure precision, outperforming existing baselines. For art style erasure, although our method does not achieve optimal results, it still delivers significant improvements over SD models. Notably, our approach consistently preserves harmless concepts more effectively.

\subsection{Parameter Analysis}
\label{sec:analysis}
We adopt quantitative metrics to evaluate erasure performance. Specifically, we assess erasure effectiveness using \textbf{ASR} and \textbf{CLIP} on 500 images generated from five prompts: \textit{'nudity', 'naked person', 'nude art', 'naked woman', 'naked man'}, and use \textbf{FID} to measure the generation quality.

\begin{figure}[t]
    \centering
    % 左侧 minipage (占据 49% 宽度)
    \begin{minipage}{0.49\linewidth}
        \centering
        \includegraphics[width=1\linewidth]{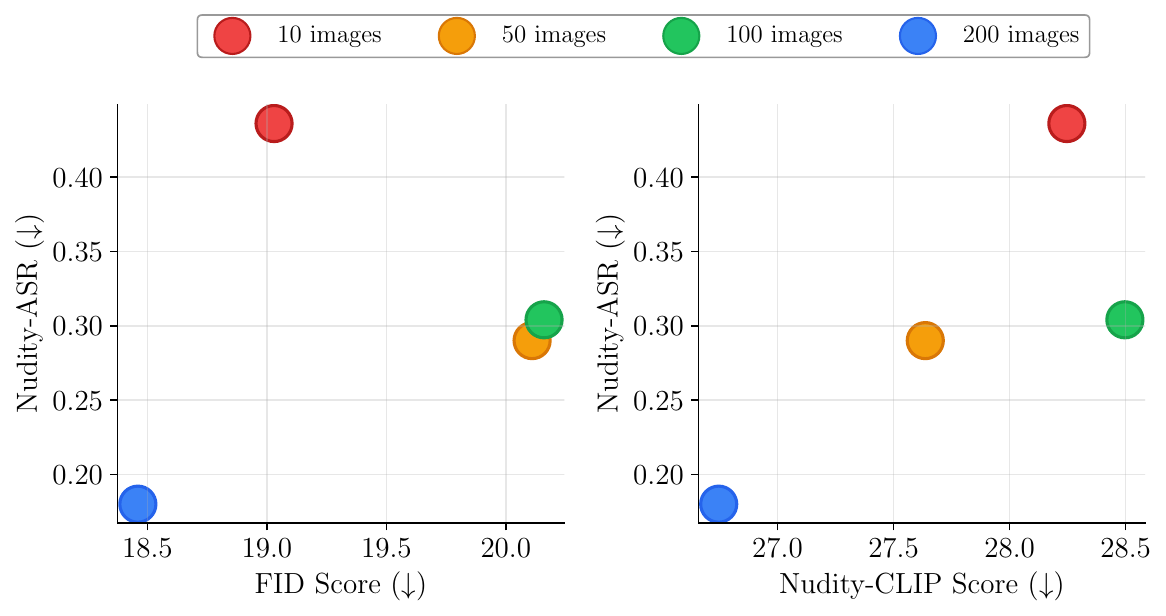}
        \caption{\textbf{Effect of training image number on erasure performance.} Performance trade-offs between FID vs. ASR and CLIP vs. ASR for different numbers of training images. Lower values indicate better performance for all metrics.}
        \label{fig:number}
    \end{minipage}
    \hfill % 插入水平间距
    \begin{minipage}{0.49\linewidth}
        \centering
        \includegraphics[width=1\linewidth]{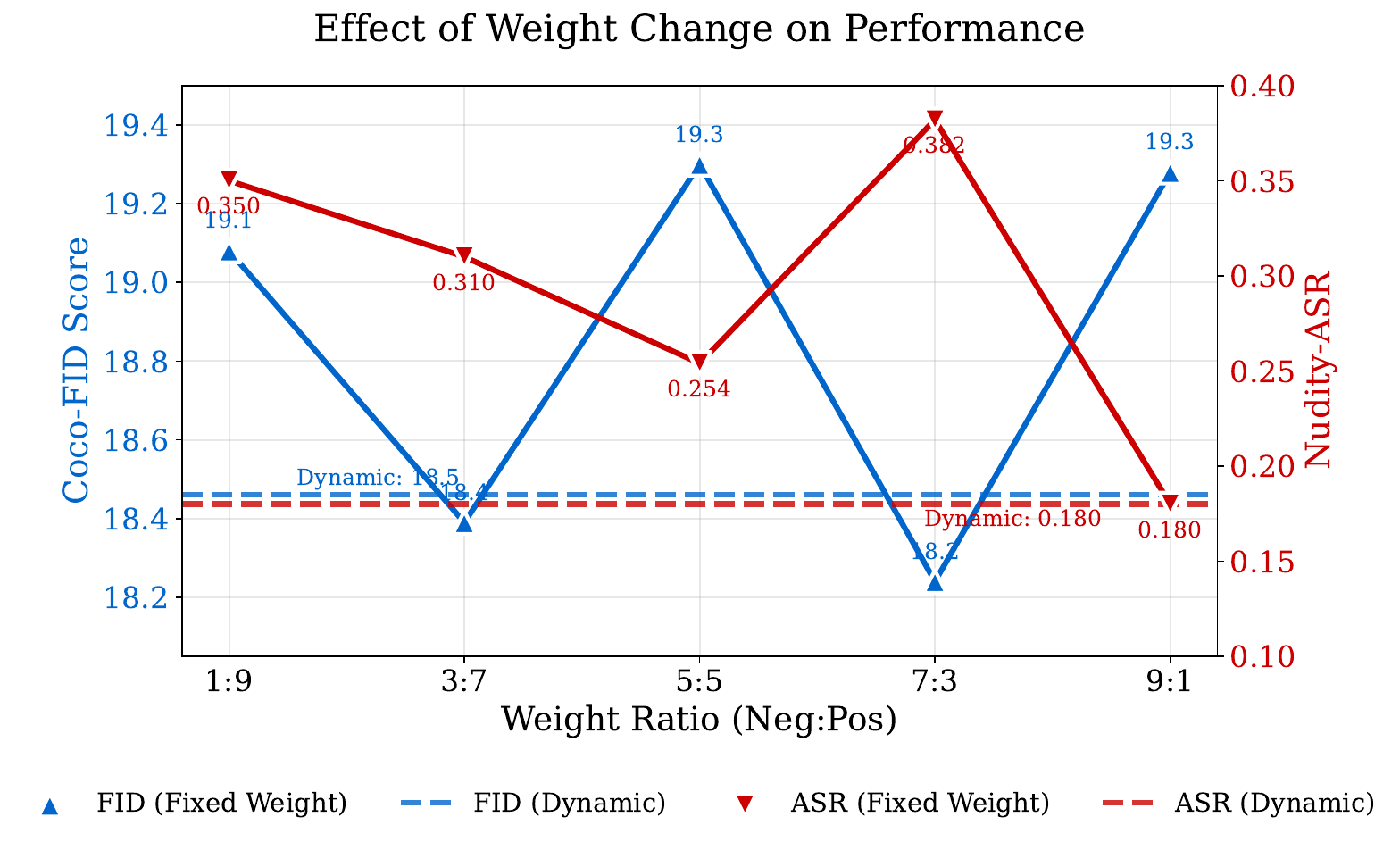}
        \caption{\textbf{Impact of Different Weights on Erasure.} Blue lines represent FID scores, red lines represent ASR. Dashed lines show dynamic weights, solid lines show fixed weights.}
        \label{fig:weight}
    \end{minipage}
    \vspace{-0.1cm}
\end{figure}

\noindent\textbf{Impact of training image number}.We train the model using varying numbers of images (10, 50, 100, 200). As shown in Figure~\ref{fig:number}, increasing the number of training images generally improves both erasure performance and generation quality. Specifically, the ASR decreases from 0.436 to 0.18 and CLIP scores improve from 28.25 to 26.75, indicating stronger erasure capability, while FID scores improve from 19.03 to 18.46, demonstrating better generation quality. We observe that performance gains are most significant when increasing from 10 to 50 images, with more modest improvements between 50 and 100 images, and continued enhancement at 200 images. Based on this analysis, we select 200 images for training as it achieves the optimal balance between erasure effectiveness and generation quality across all metrics.

\noindent\textbf{Dynamic Weight vs. Fixed Weight}. 
We explore the effects of different negative-to-positive weight ratios, comparing fixed and dynamic weighting strategies. Fixed weights maintain constant ratios (1:9, 3:7, 5:5, 7:3, 9:1), while dynamic weights adjust adaptively during training. As shown in Figure~\ref{fig:weight}, there exists a clear trade-off between generation quality and erasure effectiveness across different weight ratios. Notably, the dynamic weighting approach consistently outperforms fixed weighting, achieving better FID scores and lower ASR, demonstrating superior balance between generation quality and erasure effectiveness.

\subsection{Ablation Study}\label{sec:ablation}

\begin{wraptable}{r}{0.5\linewidth}
\centering
\footnotesize
\setlength{\tabcolsep}{5pt}
\renewcommand{\arraystretch}{1.15}
\vspace{-10pt}
\caption{\textbf{Ablation study on framework components.}}
\label{tab:ablation}
\begin{tabular}{lcccccc}
\toprule
& \multicolumn{3}{c}{\textbf{Components}} &
 \multicolumn{3}{c}{\textbf{Performance}} \\
\cmidrule(lr){2-4}\cmidrule(lr){5-7}
\textbf{Config} & Text & BI & Mask & ASR↓ & CLIP↓ & FID↓ \\
\midrule
Text       & \checkmark & & & 0.48 & 27.47 & 18.18 \\
Text + BI  & \checkmark & \checkmark & & 0.32 & 26.19 & 18.92 \\
BI         &  & \checkmark & & 0.90 & 29.24 & 17.58 \\
BI + Mask  &  & \checkmark & \checkmark & 0.86 & 29.03 & \textbf{16.97} \\
\rowcolor{gray!15}
Full Method & \checkmark & \checkmark & \checkmark & \textbf{0.18} & \textbf{25.40} & 18.46 \\
\bottomrule
\end{tabular}
\vspace{-5pt}
\end{wraptable}

We conduct an ablation study to evaluate the contribution of each component in our framework, including textual supervision (\textbf{Text}), bidirectional image guidance (\textbf{BI}), and mask-based localization (\textbf{Mask}). As shown in Table~\ref{tab:ablation}, using \textbf{Text} alone achieves an ASR of 0.48. Adding \textbf{BI} significantly reduces ASR to 0.32, demonstrating the effectiveness of bidirectional image guidance in suppressing undesired semantics. However, the \textbf{BI}-only configuration performs poorly with ASR of 0.90, indicating that textual constraints are essential for semantic grounding. The \textbf{BI + Mask} configuration shows marginal improvement over \textbf{BI} alone but still underperforms compared to text-based methods. Our \textbf{Full Method} achieves the best erasure performance with ASR of 0.18 and the lowest CLIP score of 25.40, confirming that the combination of all three components yields optimal balance between erasure efficacy and visual fidelity.

\section{Conclusion}

\noindent Current concept erasure methods primarily employ unidirectional negative concept suppression, which leads to over-erasure and semantic collapse.To address this, we introduce bidirectional image branches for training based on text-image fusion erasure work, while simultaneously utilizing mask-based images for training. Our approach effectively balances image generation quality, erasure effectiveness, and resistance to attacks. However,our work also has limitations. Identifying appropriate guiding concepts for positive guidance remains a challenge that warrants further investigation.

% \section*{Acknowledgements}
\newpage
\bibliography{iclr2026_conference}

@inproceedings{rombach2022high,
  title     = {High-resolution image synthesis with latent diffusion models},
  author    = {Rombach, Robin and Blattmann, Andreas and Lorenz, Dominik and Esser, Patrick and Ommer, Bj{\"o}rn},
  booktitle = {CVPR},
  pages     = {10684--10695},
  year      = {2022}
}

@article{tatum2023porn,
  title     = {AI porn is easy to make now. For women, that's a nightmare.},
  author    = {Hunter, Tatum},
  journal   = {The Washington Post},
  year      = {2023}
}

@inproceedings{schramowski2023safe,
  title     = {Safe latent diffusion: Mitigating inappropriate degeneration in diffusion models},
  author    = {Schramowski, Patrick and Brack, Manuel and Deiseroth, Bj{\"o}rn and Kersting, Kristian},
  booktitle = {CVPR},
  pages     = {22522--22531},
  year      = {2023}
}

@inproceedings{jiang2023ai,
  title     = {AI Art and its Impact on Artists},
  author    = {Jiang, Harry H and Brown, Lauren and Cheng, Jessica and Khan, Mehtab and Gupta, Abhishek and Workman, Deja and Hanna, Alex and Flowers, Johnathan and Gebru, Timnit},
  booktitle = {AIES},
  pages     = {363--374},
  year      = {2023}
}

@inproceedings{DDPM,
  title     = {Denoising Diffusion Probabilistic Models},
  author    = {Ho, Jonathan and Jain, Ajay and Abbeel, Pieter},
  booktitle = {NeurIPS},
  pages     = {6840--6851},
  year      = {2020}
}

@inproceedings{CLIP,
  title     = {Learning Transferable Visual Models from Natural Language Supervision},
  author    = {Radford, Alec and Kim, Jong Wook and Hallacy, Chris and Ramesh, Aditya and Goh, Gabriel and Agarwal, Sandhini and Sastry, Girish and Askell, Amanda and Mishkin, Pamela and Clark, Jack and Krueger, Gretchen and Sutskever, Ilya},
  booktitle = {ICML},
  pages     = {8748--8763},
  year      = {2021}
}

@inproceedings{LDM,
  title     = {High-Resolution Image Synthesis with Latent Diffusion Models},
  author    = {Rombach, Robin and Blattmann, Andreas and Lorenz, Dominik and Esser, Patrick and Ommer, Bj\"{o}rn},
  booktitle = {CVPR},
  pages     = {10684--10695},
  year      = {2022}
}

@article{DDIM,
  title     = {Denoising diffusion implicit models},
  author    = {Song, Jiaming and Meng, Chenlin and Ermon, Stefano},
  journal   = {arXiv preprint arXiv:2010.02502},
  year      = {2020}
}

@inproceedings{LAION5B,
  title     = {LAION-5B: An Open Large-Scale Dataset for Training Next Generation Image-Text Models},
  author    = {Schuhmann, Christoph and Beaumont, Romain and Vencu, Richard and Gordon, Cade and Wightman, Ross and Cherti, Mehdi and Coombes, Theo and Katta, Aaron and Mullis, Clayton and Wortsman, Mitchell and others},
  booktitle = {NeurIPS},
  volume    = {35},
  pages     = {25278--25294},
  year      = {2022}
}

@misc{coyo700m,
  title     = {COYO-700M: Image-Text Pair Dataset},
  author    = {Byeon, Minwoo and Park, Beomhee and Kim, Haecheon and Lee, Sungjun and Baek, Woonhyuk and Kim, Saehoon},
  note      = {Technical Report},
  year      = {2022}
}

@inproceedings{Casanova_2021,
  title     = {Instance-Conditioned GAN},
  author    = {Casanova, Arantxa and Careil, Marlene and Verbeek, Jakob and Drozdzal, Michal and Romero-Soriano, Adriana},
  booktitle = {NeurIPS},
  pages     = {27517--27529},
  year      = {2021}
}

@inproceedings{Gao_2024,
  title     = {Eraseanything: Enabling concept erasure in rectified flow transformers},
  author    = {Gao, Daiheng and Lu, Shilin and Zhou, Wenbo and Chu, Jiaming and Zhang, Jie and Jia, Mengxi and Zhang, Bang and Fan, Zhaoxin and Zhang, Weiming},
  booktitle = {ICML},
  year      = {2025}
}

@article{Li_2023,
  title     = {Diffusion models for image restoration and enhancement: a comprehensive survey},
  author    = {Li, Xin and Ren, Yulin and Jin, Xin and Lan, Cuiling and Wang, Xingrui and Zeng, Wenjun and Wang, Xinchao and Chen, Zhibo},
  journal   = {International Journal of Computer Vision},
  pages     = {1--31},
  year      = {2025}
}

@inproceedings{Ramesh_2021,
  title     = {Zero-Shot Text-to-Image Generation},
  author    = {Ramesh, Aditya and Pavlov, Mikhail and Goh, Gabriel and Gray, Scott and Voss, Chelsea and Radford, Alec and Chen, Mark and Sutskever, Ilya},
  booktitle = {ICML},
  pages     = {8821--8831},
  year      = {2021}
}

@inproceedings{Unet,
  title     = {U-Net: Convolutional Networks for Biomedical Image Segmentation},
  author    = {Ronneberger, Olaf and Fischer, Philipp and Brox, Thomas},
  booktitle = {MICCAI},
  pages     = {234--241},
  year      = {2015}
}

@inproceedings{Takagi_2023,
  title     = {High-Resolution Image Reconstruction with Latent Diffusion Models from Human Brain Activity},
  author    = {Takagi, Yu and Nishimoto, Shinji},
  booktitle = {CVPR},
  pages     = {14453--14463},
  year      = {2023}
}

@misc{Walton_2022,
  title     = {StyleNat: Giving Each Head a New Perspective},
  author    = {Walton, Steven and Hassani, Ali and Xu, Xingqian and Wang, Zhangyang and Shi, Humphrey},
  note      = {arXiv:2206.15515},
  year      = {2022}
}

@inproceedings{Wu_Reconfusion_2024,
  title     = {Reconfusion: 3D Reconstruction with Diffusion Priors},
  author    = {Wu, Rundi and Mildenhall, Ben and Henzler, Philipp and Park, Keunhong and Gao, Ruiqi and Watson, Daniel and Srinivasan, Pratul P. and Verbin, Dor and Barron, Jonathan T. and Poole, Ben and others},
  booktitle = {CVPR},
  pages     = {21551--21561},
  year      = {2024}
}

@misc{Yu_2022,
  title     = {Scaling Autoregressive Models for Content-Rich Text-to-Image Generation},
  author    = {Yu, Jiahui and Xu, Yuanzhong and Koh, JingYu and Luong, Thang and Baid, Gunjan and Wang, Zirui and Vasudevan, Vijay and Ku, Alexander and Yang, Yinfei and Ayan, Burcu Karagol and Hutchinson, Ben and Han, Wei and Parekh, Zarana and Li, Xin and Zhang, Han and Baldridge, Jason and Wu, Yonghui},
  note      = {arXiv:2208.14788},
  year      = {2022}
}

@inproceedings{Zhang_2023,
  title     = {Adding Conditional Control to Text-to-Image Diffusion Models},
  author    = {Zhang, Lvmin and Rao, Anyi and Agrawala, Maneesh},
  booktitle = {ICCV},
  pages     = {3836--3847},
  year      = {2023}
}

@inproceedings{Zhou_2023,
  title     = {Shifted Diffusion for Text-to-Image Generation},
  author    = {Zhou, Yufan and Liu, Bingchen and Zhu, Yizhe and Yang, Xiao and Chen, Changyou and Xu, Jinhui},
  booktitle = {CVPR},
  pages     = {10157--10166},
  year      = {2023}
}

@inproceedings{FID,
  title     = {GANs Trained by a Two Time-Scale Update Rule Converge to a Local Nash Equilibrium},
  author    = {Heusel, Martin and Ramsauer, Hubert and Unterthiner, Thomas and Nessler, Bernhard and Hochreiter, Sepp},
  booktitle = {NeurIPS},
  pages     = {6626--6637},
  year      = {2017}
}

@inproceedings{ESD,
  title     = {Erasing Concepts from Diffusion Models},
  author    = {Gandikota, Rohit and Materzynska, Joanna and Fiotto-Kaufman, Jaden and Bau, David},
  booktitle = {ICCV},
  pages     = {2426--2436},
  year      = {2023}
}

@inproceedings{Coerasing,
  title     = {One Image is Worth a Thousand Words: A Usability Preservable Text-Image Collaborative Erasing Framework},
  author    = {Li, Feiran and Xu, Qianqian and Bao, Shilong and Yang, Zhiyong and Cao, Xiaochun and Huang, Qingming},
  booktitle = {ICML},
  year      = {2025}
}

@inproceedings{TRCE,
  title     = {TRCE: Towards Reliable Malicious Concept Erasure in Text-to-Image Diffusion Models},
  author    = {Chen, Ruidong and Guo, Honglin and Wang, Lanjun and Zhang, Chenyu and Nie, Weizhi and Liu, An-An},
  booktitle = {ICCV},
  year      = {2025}
}

@inproceedings{RACE,
  title     = {R.A.C.E.: Robust Adversarial Concept Erasure for Secure Text-to-Image Diffusion Model},
  author    = {Kim, Changhoon and Min, Kyle and Yang, Yezhou},
  booktitle = {ECCV},
  pages     = {461--478},
  year      = {2024}
}

@inproceedings{RECE,
  title     = {Reliable and Efficient Concept Erasure of Text-to-Image Diffusion Models},
  author    = {Chao, Gong and Kai, Chen and Zhipeng, Wei and Jingjing, Chen and Yu-Gang, Jiang},
  booktitle = {ECCV},
  pages     = {73--88},
  year      = {2024}
}

@inproceedings{SalUn,
  title     = {SalUn: Empowering Machine Unlearning via Gradient-Based Weight Saliency in Both Image Classification and Generation},
  author    = {Fan, Chongyu and Liu, Jiancheng and Zhang, Yihua and Wong, Eric and Wei, Dennis and Liu, Sijia},
  booktitle = {ICLR},
  year      = {2024}
}

@inproceedings{MACE,
  title     = {MACE: Mass Concept Erasure in Diffusion Models},
  author    = {Lu, Shilin and Wang, Zilan and Li, Leyang and Liu, Yanzhu and Kong, Adams Wai-Kin},
  booktitle = {CVPR},
  pages     = {6430--6440},
  year      = {2024}
}

@inproceedings{Scissor,
  title     = {Scissorhands: Scrub data influence via connection sensitivity in networks},
  author    = {Wu, Jing and Harandi, Mehrtash},
  booktitle = {ECCV},
  pages     = {367--384},
  year      = {2024}
}

@inproceedings{FMN,
  title     = {Forget-me-not: Learning to forget in text-to-image diffusion models},
  author    = {Zhang, Gong and Wang, Kai and Xu, Xingqian and Wang, Zhangyang and Shi, Humphrey},
  booktitle = {CVPR},
  pages     = {1755--1764},
  year      = {2024}
}

@inproceedings{AdvUnlearn,
  title     = {Defensive Unlearning with Adversarial Training for Robust Concept Erasure in Diffusion Models},
  author    = {Zhang, Yimeng and Chen, Xin and Jia, Jinghan and Zhang, Yihua and Fan, Chongyu and Liu, Jiancheng and Hong, Mingyi and Ding, Ke and Liu, Sijia},
  booktitle = {NeurIPS},
  pages     = {36748--36776},
  year      = {2024}
}

@inproceedings{SPM,
  title     = {One-Dimensional Adapter to Rule Them All: Concepts, Diffusion Models and Erasing Applications},
  author    = {Lyu, Mengyao and Yang, Yuhong and Hong, Haiwen and Chen, Hui and Jin, Xuan and He, Yuan and Xue, Hui and Han, Jungong and Ding, Guiguang},
  booktitle = {CVPR},
  pages     = {7559--7568},
  year      = {2024}
}

@inproceedings{UnlearnDiffAtk,
  title     = {To Generate or Not? Safety-Driven Unlearned Diffusion Models Are Still Easy to Generate Unsafe Images...For Now},
  author    = {Zhang, Yimeng and Chen, Xin and Chen, Aochuan and Zhang, Yihua and Liu, Jiancheng and Ding, Ke},
  booktitle = {ECCV},
  pages     = {385--403},
  year      = {2024}
}

@inproceedings{UCE,
  title     = {Unified Concept Editing in Diffusion Models},
  author    = {Gandikota, Rohit and Orgad, Hadas and Belinkov, Yonatan and Materzy{\'n}ska, Joanna and Bau, David},
  booktitle = {WACV},
  pages     = {5111--5120},
  year      = {2024}
}
\bibliographystyle{colm2025_conference}

\newpage
\appendix
\renewcommand \thepart{} % make "Part" text invisible
    \renewcommand \partname{}
\part{Appendix} % Start the appendix part

    \parttoc % Insert the appendix TOC
%\clearpage
%\setcounter{page}{1}
%\maketitlesupplementary

\section{Limitations}
Our approach effectively balances image generation quality, erasure effectiveness, and resistance to attacks. However, our work also has limitations. Identifying appropriate guiding concepts for positive guidance remains a challenge that warrants further investigation.

\begin{center}
    % 0.5\linewidth 表示占当前行宽的 50%，你可以改成 0.6 或 0.7 调整宽度
    \begin{minipage}{0.6\linewidth} 
        \begin{algorithm}[H] % <--- 必须用 [H]，强制固定在 minipage 盒子里
            \small
            \caption{Bidirectional Image Guidance Training}
            \label{alg:biig}
            \begin{algorithmic}[1]
                \Require $\theta_0, \theta, E_{\mathrm{img}}, P, \mathcal{D}_{\text{neg}}, \mathcal{D}_{\text{pos}}, \{M\}$
                \For{$\text{iter} = 1 \to T_{\text{max}}$}
                    \State Sample $I_{\text{neg}} \sim \mathcal{D}_{\text{neg}}$, $I_{\text{pos}} \sim \mathcal{D}_{\text{pos}}$
                    \State Apply masks via Eq.\,\eqref{eq:mask_preprocess}: $\tilde{I}_{\text{neg}}, \tilde{I}_{\text{pos}}$
                    \State Encode: $e_{\text{neg}} = E_{\mathrm{img}}(T(\tilde{I}_{\text{neg}}))$, $e_{\text{pos}} = E_{\mathrm{img}}(T(\tilde{I}_{\text{pos}}))$
                    \State Project: $c_{\text{neg}} = P(e_{\text{neg}})$, $c_{\text{pos}} = P(e_{\text{pos}})$
                    \State Sample $t, Z_t$; Query $\theta_0$: $\epsilon_u^{\mathrm{ref}}, \epsilon_{\text{neg}}^{\mathrm{ref}}, \epsilon_{\text{pos}}^{\mathrm{ref}}$
                    \State Build targets: $\epsilon_{\mathrm{tgt}}^{\text{neg}}, \epsilon_{\mathrm{tgt}}^{\text{pos}}$
                    \State Predict: $\hat{\epsilon}_{\text{neg}}, \hat{\epsilon}_{\text{pos}}$ via Eq.\,\eqref{eq:noice}
                    \State Compute: $\mathcal{L}_{\text{neg}}, \mathcal{L}_{\text{pos}}$ via Eq.\,\eqref{eq:compute}
                    \State $\mathcal{L} = \lambda_{\text{neg}}\mathcal{L}_{\text{neg}} + \lambda_{\text{pos}}\mathcal{L}_{\text{pos}}$
                    \State Update $\theta$ via $\nabla_\theta \mathcal{L}$
                \EndFor
            \end{algorithmic}
        \end{algorithm}
    \end{minipage}
\end{center}

\begin{figure}[t]
\centering
\includegraphics[width=1\linewidth]{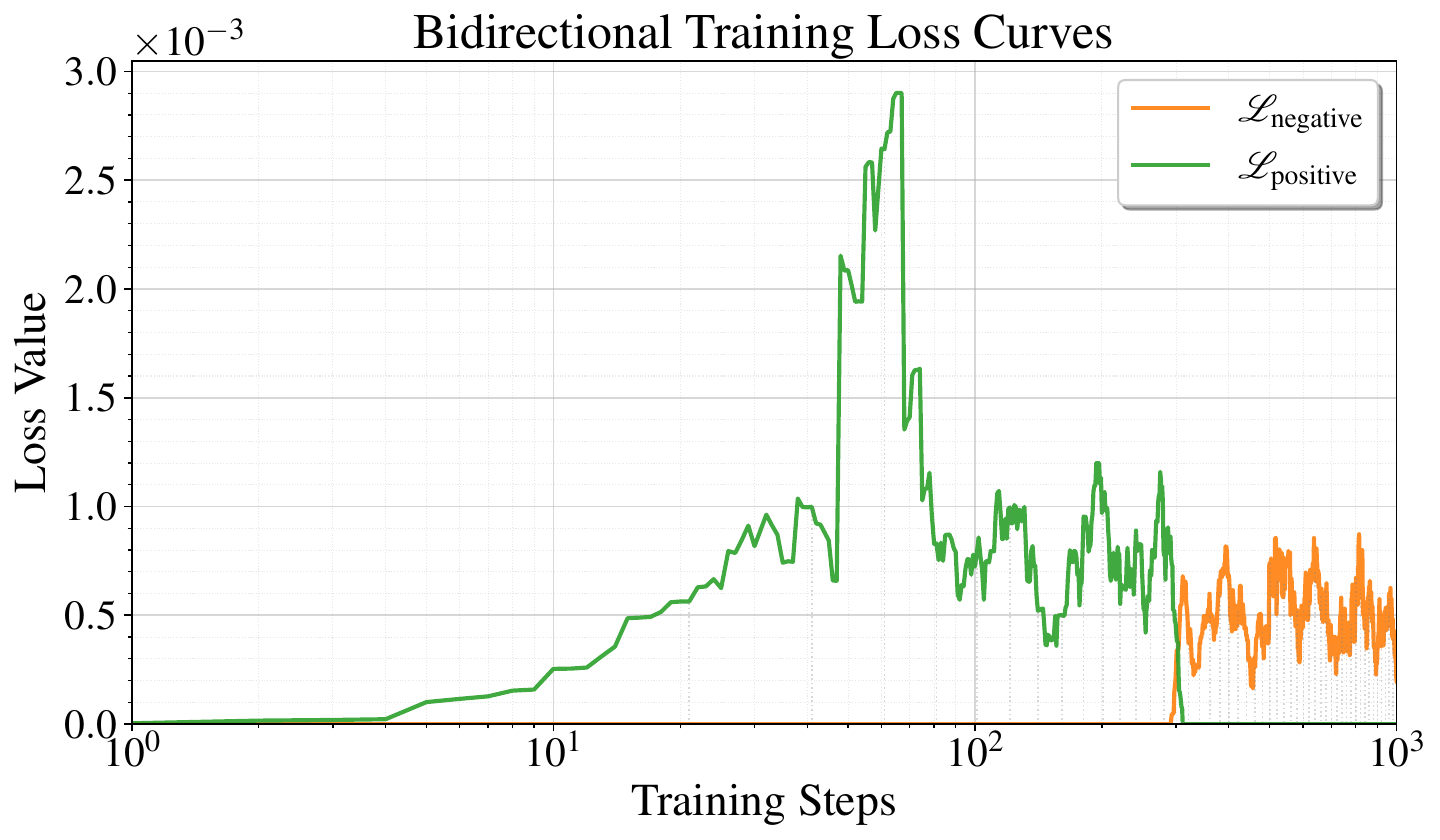}
\caption{\textbf{Loss Changes During Training.} Training with dynamic weight changes: green indicates positive guidance, orange indicates negative guidance, with the horizontal axis exhibiting logarithmic scale.}
\label{fig:loss}
\end{figure}

\begin{figure}[t]
\centering
\includegraphics[width=0.9\linewidth]{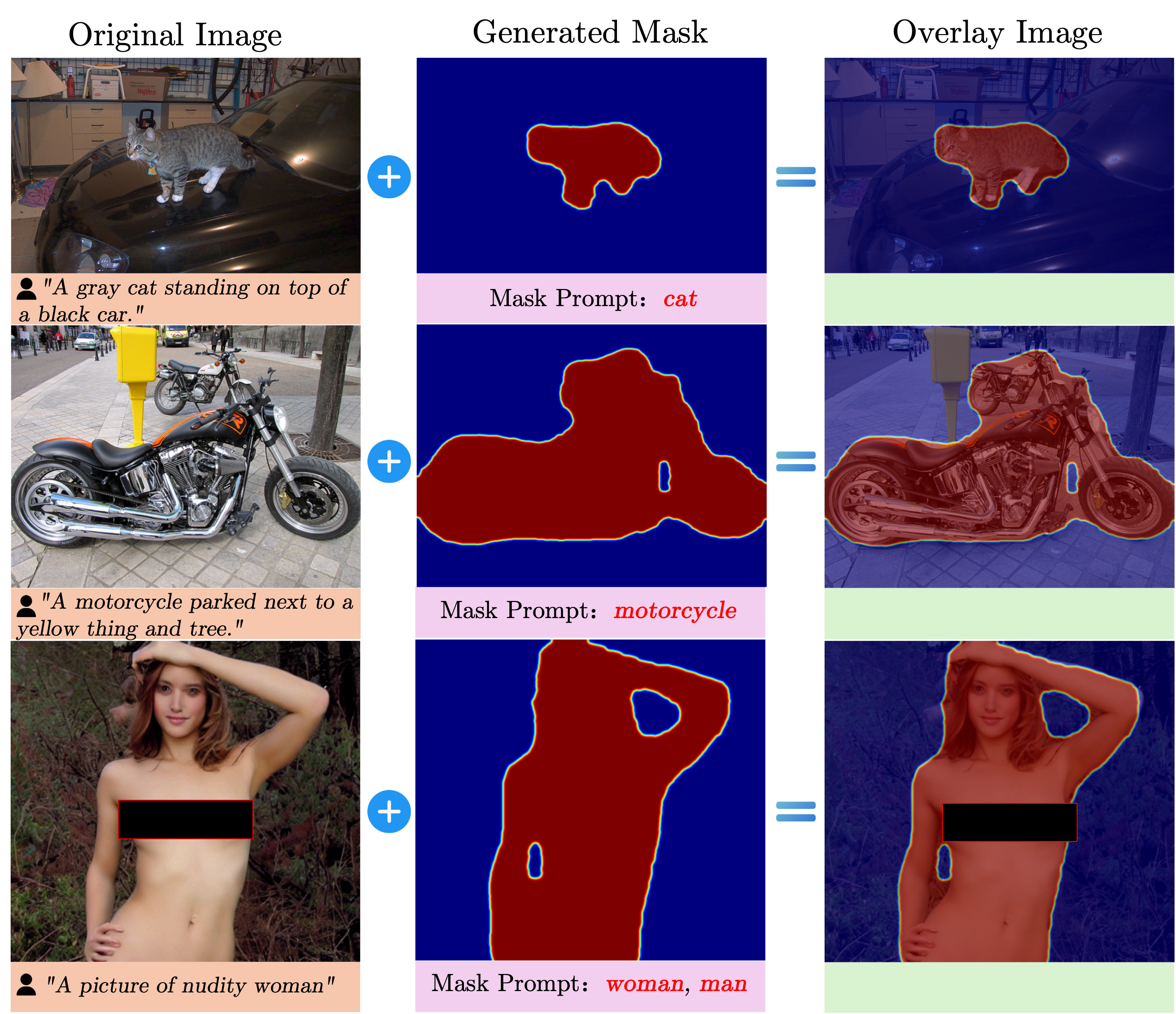}
\caption{\textbf{Mask-guided image preprocessing.} (\textit{left}) the original image, (\textit{middle}) the generated semantic mask, and (\textit{right}) the masked image. The mask highlights target regions while suppressing background clutter.}
\label{fig:mask}
\end{figure}

\section{Experimental Details}
We generate images using an NVIDIA RTX 4090 with 64-bit precision and 50 inference steps, requiring 4 seconds per image. Training our model simultaneously requires two RTX 4090 GPUs with a learning rate set to $10^{-5}$, taking 40 minutes to converge, as shown in Figure~\ref{fig:loss}. As the number of training steps increases, the model's training loss first rises and then decreases. It achieves basic convergence during the forward guidance phase and fluctuates during the backward guidance phase.

\vspace{0.6em}
\noindent\textbf{Evaluation Details.} 

% \vspace{0.6em}
% \noindent For evaluating nudity-related concepts, we employ the NudeNet detector with the following categories: \texttt{FEMALE\_GENITALIA\_EXPOSED},\texttt{ANUS\_EXPOSED},
% \texttt{FEMALE\_BREAST\_EXPOSED},\texttt{BUTTOCKS\_EXPOSED},
% \texttt{MALE\_BREAST\_EXPOSED},\texttt{MALE\_GENITALIA\_EXPOSED}, using a detection threshold of 0.5. 

\vspace{0.6em}
\noindent For evaluating nudity-related concepts, we employ the NudeNet detector with the following categories: 
\texttt{MALE\_BREAST\_EXPOSED},\texttt{MALE\_GENITALIA\_EXPOSED},
\texttt{FEMALE\_BREAST\_EXPOSED},\texttt{BUTTOCKS\_EXPOSED},
\texttt{FEMALE\_GENITALIA\_EXPOSED},\texttt{ANUS\_EXPOSED},using a detection threshold of 0.5. 

\vspace{0.6em}
\noindent For evaluation on the COCO dataset, we utilize the cleanFID library to compute the FID metric.

\vspace{0.6em}
\noindent For celebrity concept erasure, we employ a GCD evaluator using OpenCV's DNN face detector with a face margin of 0.2, selecting the most probable match. Evaluation utilizes five distinct prompt templates: ``A portrait of [Celebrity Name]'', ``An image capturing [Celebrity Name] at a public event'', ``An oil painting of [Celebrity Name]'', ``A sketch of [Celebrity Name]'', and ``[Celebrity Name] in an official photo''. Each template employs multiple random seeds to ensure robust and statistically significant evaluation.

\begin{figure}[t]
\centering
\includegraphics[width=1\linewidth]{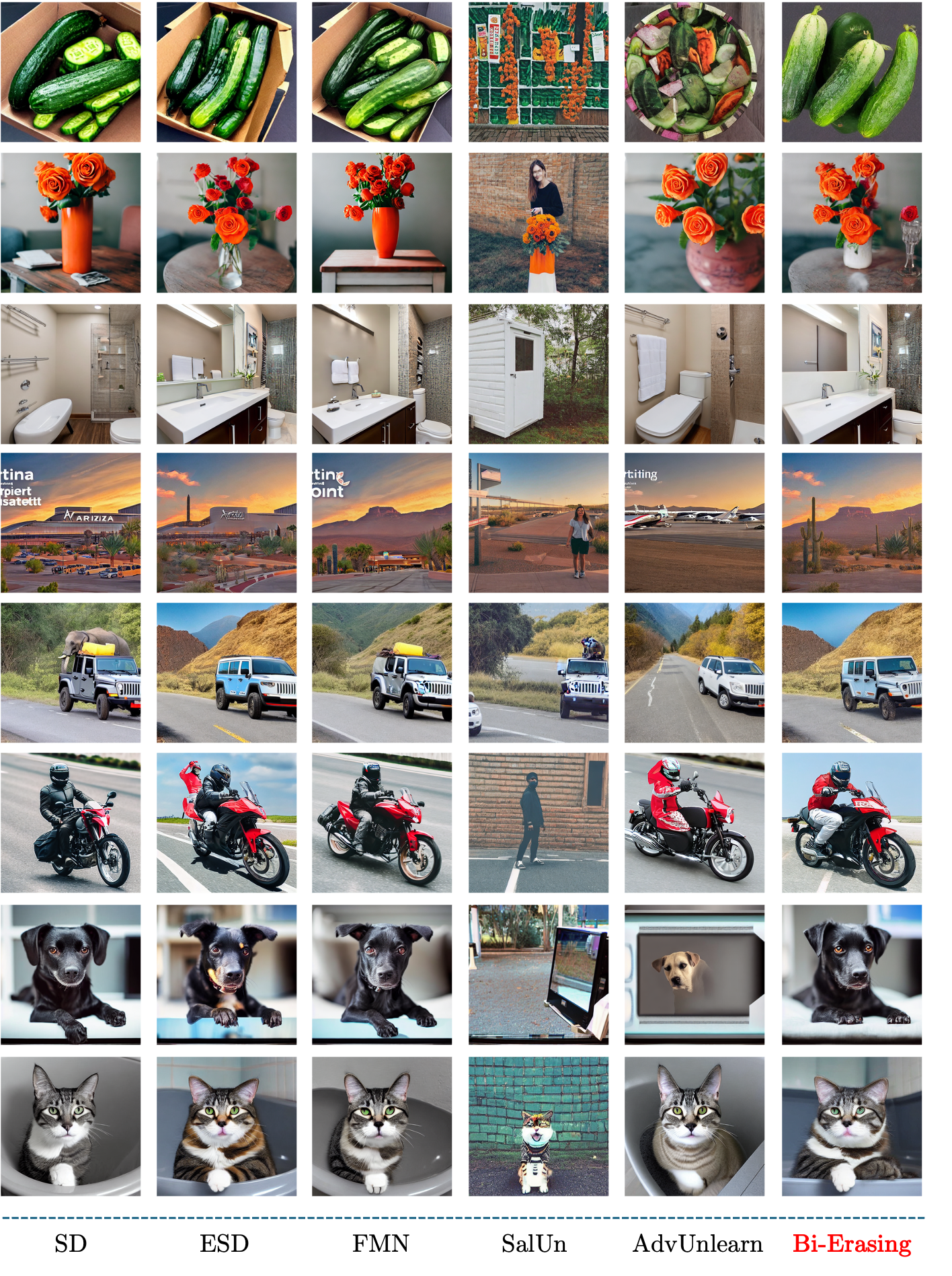}
\caption{\textbf{Images generated from the COCO dataset using different erasure methods.} Each column represents a method, and each row represents the same prompt.}
\label{fig:FID}
\end{figure}

\section{Results Display}

\vspace{0.6em}
\noindent\textbf{COCO Dataset Image Generation.} 
As shown in Figure~\ref{fig:FID}, we present a selection of images generated from the COCO dataset under different erasure strategies to demonstrate the impact of erasure methods on generation quality. It can be observed that most erasure methods effectively preserve the generation quality of unrelated concepts, but the SalUn method yields poor generation quality.

\begin{figure}[t]
\centering
\includegraphics[width=1\linewidth]{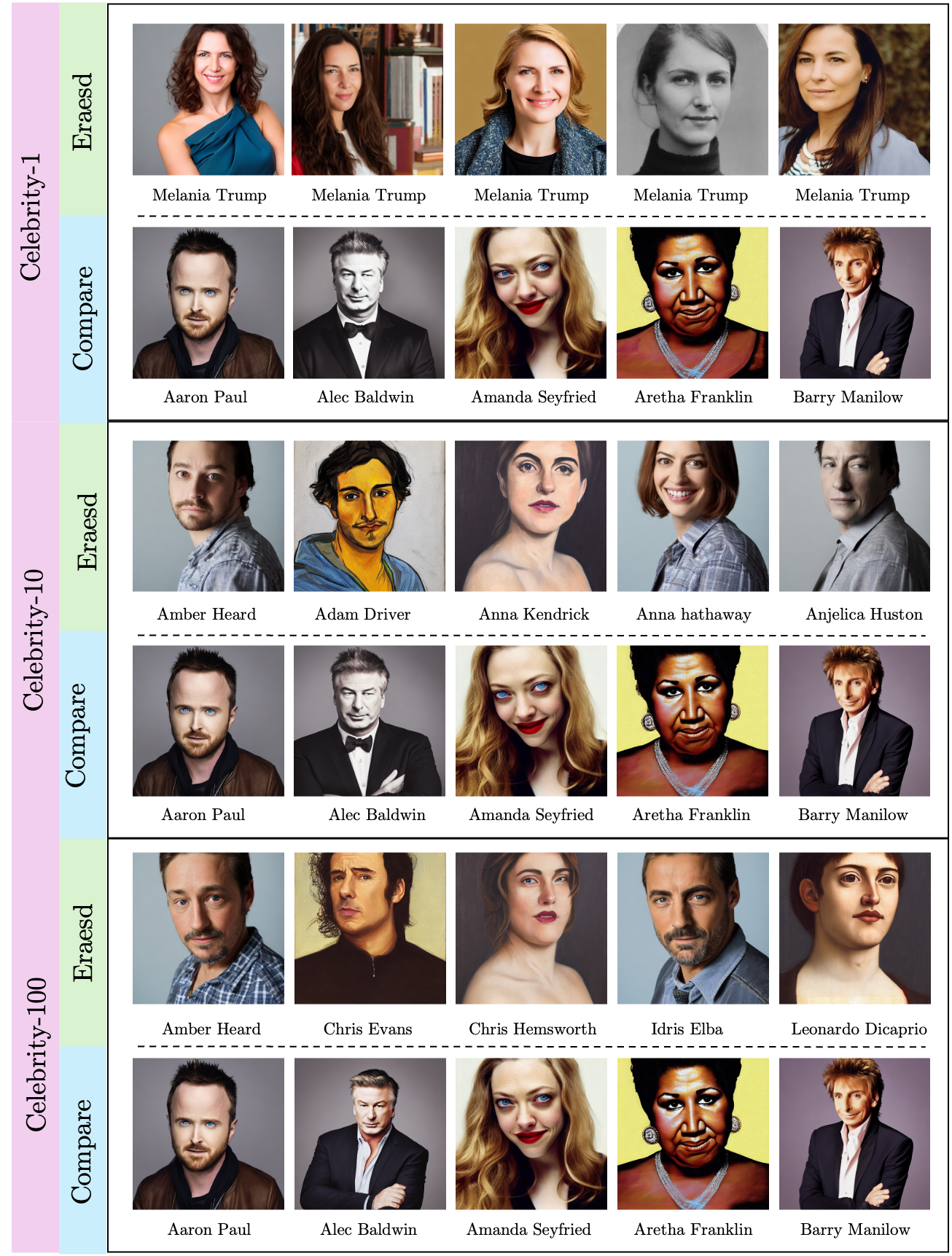}
\caption{\textbf{The BiErasing method acts upon the MACE method to perform multi-concept celebrity erasure.} Each quantity of celebrity erasure is demonstrated using a control group and an erasure group.}
\label{fig:mace+bi}
\end{figure}

\noindent\textbf{Celebrity Concept Erasure.} 
We demonstrate the celebrity erasure results for varying quantities in Figure~\ref{fig:mace+bi}, ranging from 1 to 10 to 100 celebrities. For each scale, both erasure and control groups are presented. It can be observed that our method effectively erases the target celebrity concept while preserving the generation of unrelated celebrities.

\begin{figure}[t]
    \centering
    \includegraphics[width=1\linewidth]{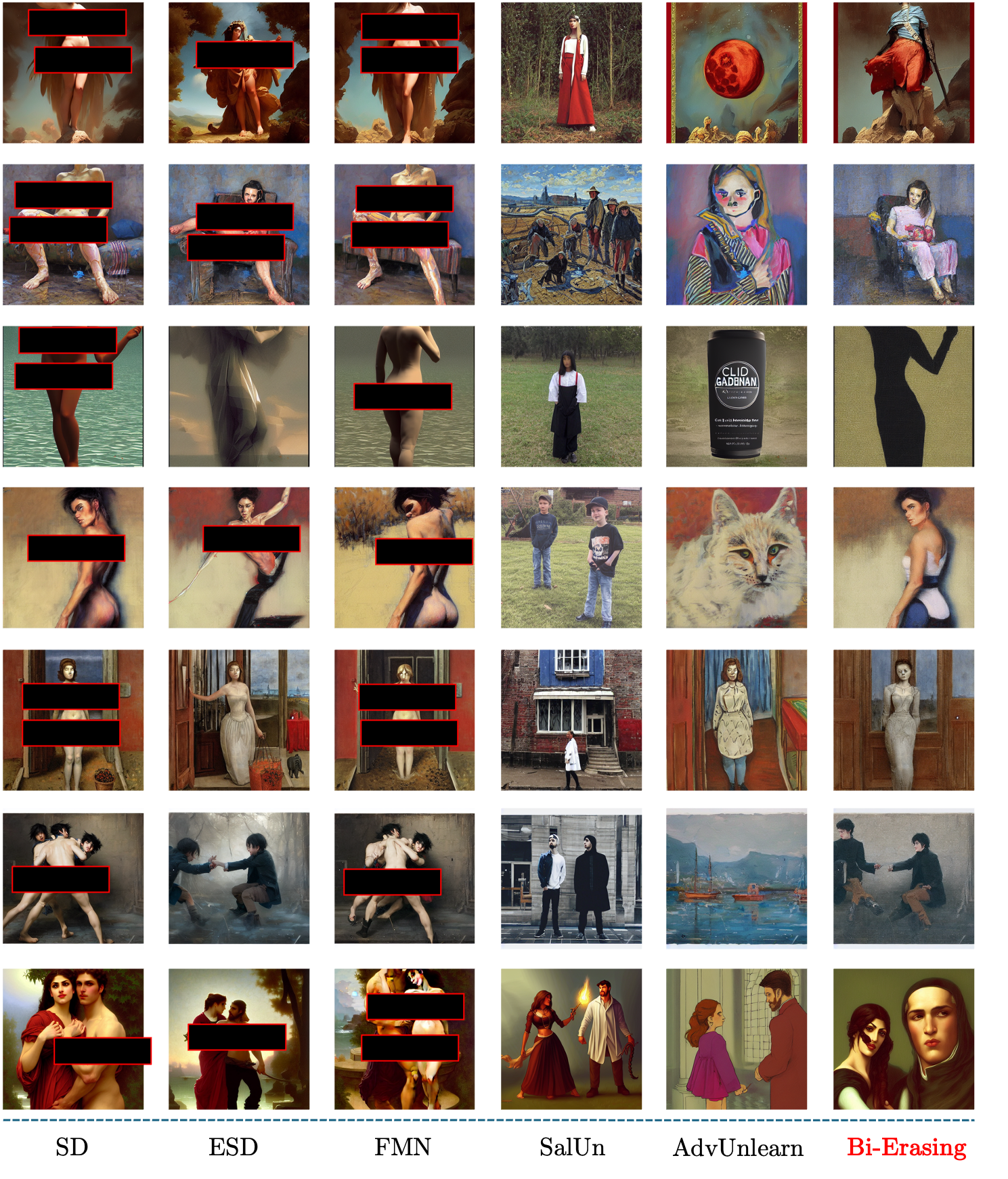}
    \caption{\textbf{Demonstration of Different Concept Erasure Methods Under Attack (Part I).} Results showing the first set of prompts.}
    \label{fig:attack_part1}
\end{figure}

\begin{figure}[t]
    \centering
    \includegraphics[width=1\linewidth]{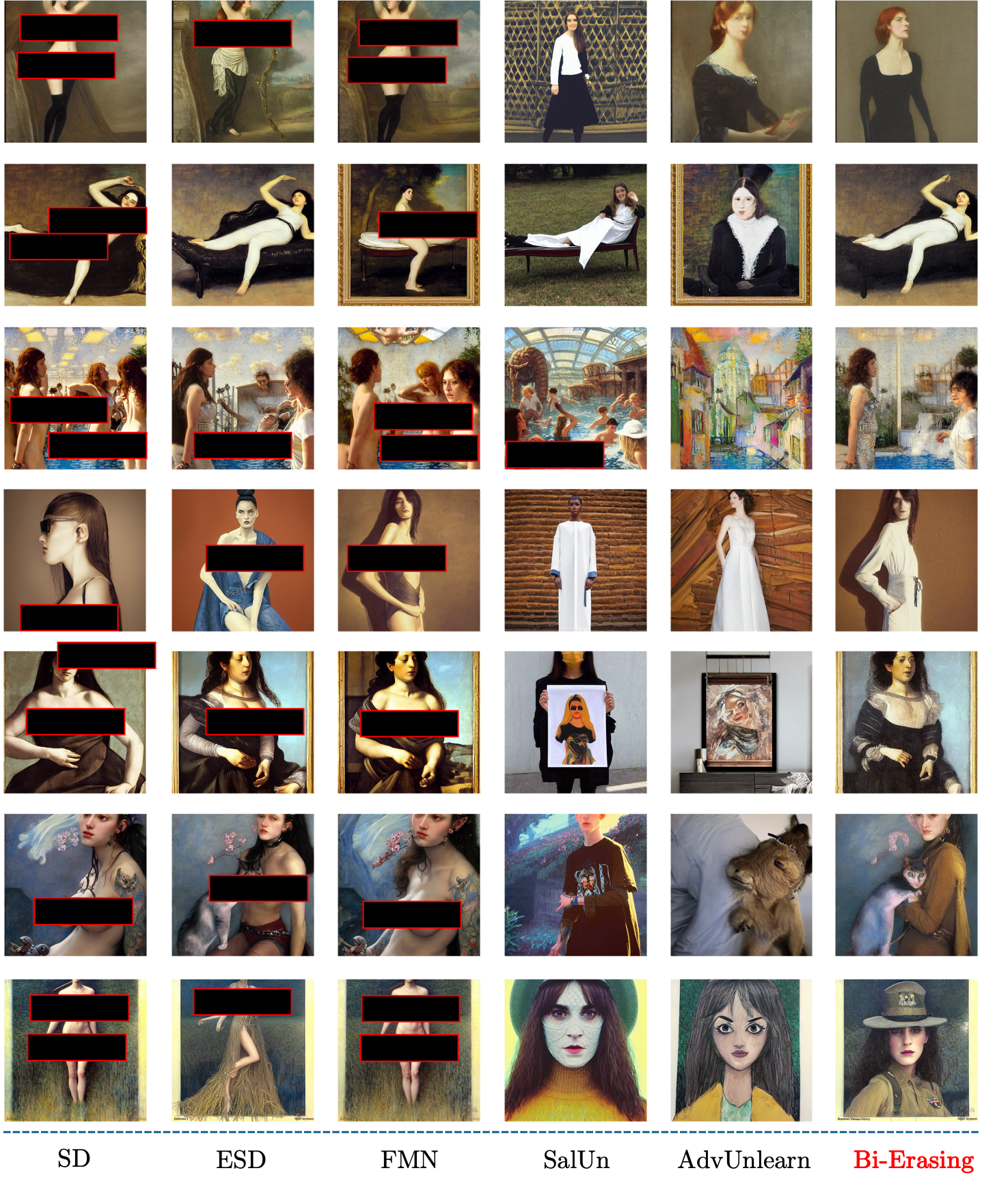}
    \caption{\textbf{Demonstration of Different Concept Erasure Methods Under Attack (Part II).} Results showing the second set of prompts.}
    \label{fig:attack_part2}
\end{figure}

\noindent\textbf{Countering Attacks.} 
In Figures~\ref{fig:attack_part1} and \ref{fig:attack_part2}, we present supplementary images demonstrating how different methods produce varying image generation outcomes when subjected to UnlearnDiffAtk attacks. Our approach typically maintains consistent style and avoids generating nude images under attack. While some competing methods resist attacks, the generated images often exhibit style shifts unrelated to the prompt.

\end{document}